\documentclass[Afour,sagev,times]{sagej}
\usepackage{moreverb,url}
\usepackage{tikz} 
\usepackage{listings} 
\usepackage{algorithm}  
\usepackage{algpseudocode}  
\usepackage{caption}  
\usepackage{xcolor} 
\usepackage{subcaption}
\usepackage{pgffor}
\usepackage{graphicx}
\usepackage{enumitem}
\usepackage{natbib}
\usepackage{array}
\usepackage{multirow}
\usepackage{makecell}   
\usepackage[colorlinks,bookmarksopen,bookmarksnumbered,citecolor=red,urlcolor=red]{hyperref}

\newcommand\BibTeX{{\rmfamily B\kern-.05em \textsc{i\kern-.025em b}\kern-.08em
T\kern-.1667em\lower.7ex\hbox{E}\kern-.125emX}}

\begin{document}

\runninghead{Bitboard version of Tetris AI}

\title{Bitboard version of Tetris AI}

\author{Xingguo Chen, Pingshou Xiong,  Zhenyu Luo, Mengfei Hu, Xinwen Li, Yongzhou Lü, Guang Yang, Chao Li, Shangdong Yang}

\corrauth{Xingguo Chen, School of Computer Science,
Nanjing University of Posts and Telecommunications,
Nanjing, Jiangsu, 210023, China}

\email{chenxg@njupt.edu.cn}

\begin{abstract}
The efficiency of game engines and policy optimization algorithms is crucial for training reinforcement learning (RL) agents in complex sequential decision-making tasks, such as Tetris. Existing Tetris implementations suffer from low simulation speeds, suboptimal state evaluation, and inefficient training paradigms, limiting their utility for large-scale RL research. To address these limitations, this paper proposes a high-performance Tetris AI framework based on bitboard optimization and improved RL algorithms. First, we redesign the Tetris game board and tetrominoes using bitboard representations, leveraging bitwise operations to accelerate core processes (e.g., collision detection, line clearing, and Dellacherie-Thiery Features extraction) and achieve a 53-fold speedup compared to OpenAI Gym-Tetris. Second, we introduce an afterstate-evaluating actor network that simplifies state value estimation by leveraging Tetris’s afterstate property, outperforming traditional action-value networks with fewer parameters. Third, we propose a buffer-optimized Proximal Policy Optimization (PPO) algorithm that balances sampling and update efficiency, achieving an average score of 3,829 on $10\times10$ grids within 3 minutes. Additionally, we develop a Python-Java interface compliant with the OpenAI Gym standard, enabling seamless integration with modern RL frameworks. Experimental results demonstrate that our framework enhances Tetris’s utility as an RL benchmark by bridging low-level bitboard optimizations with high-level AI strategies, providing a sample-efficient and computationally lightweight solution for scalable sequential decision-making research. 
\end{abstract}

\keywords{Reinforcement learning; Bitboard; Tetris; DT Features; Proximal Policy Optimization; Afterstate}

\maketitle

\section{Introduction}
Tetris, a classic sequential decision-making game invented by Alexey Pajitnov in 1984, has long served as a foundational benchmark for reinforcement learning (RL), approximate dynamic programming, and optimization algorithms—owing to its unique combination of challenges: an NP-hard optimal strategy problem \cite{demaine2003tetris}, an enormous state space (approximately \(7 \times 2^{200}\) configurations for $10\times20$ grids \cite{thiery2009building}), and the inevitability of game termination under adversarial piece sequences (e.g., prolonged Z/S pieces) \cite{burgiel1997lose, brzustowski1992can}. Its status is reinforced by its prominent role in the 2008-2009 Reinforcement Learning Competitions of  ICML workshops \cite{rlcompetition}, collectively solidifying Tetris’s value for testing RL agents’ ability to handle uncertainty and long-term planning.

\begin{table*}[!htb]
	\centering
	\caption{ Records in 10$\times$10 Tetris AI.}
	\label{tab:tetris_ai_records}
	\resizebox{0.98\textwidth}{!}{
		\begin{tabular}{lcccc}
			\toprule
			Method & Avg. Removed Lines  &Training Samples & Feature & Generator \\
			\midrule
			BCTS~\cite{thiery2009building} &3,000  & $6.5\times10^7$ &DT features &Random \\
			CBMPI~\cite{gabillon2013approximate,scherrer2015approximate} &
			4,300  & $8\times10^6$ & DT features + RBF heights& Random\\
			dSiLU-TD($\lambda$)~\cite{elfwing2018sigmoid} &\textbf{4,900}  & 200,000 & 
			Bertsekas features + neural networks & Random\\
			STEW~\cite{lichtenberg2019regularization} &
			4,800  & 238,000 & 
			DT features& 7-Bag random \\
			\textbf{Buffered PPO (Ours)} &
			3,850  & \textbf{61,440} & 
			DT features &Random\\
			\bottomrule
	\end{tabular}}
\end{table*}

Subsequent Tetris AI research, systematically summarized in Table \ref{tab:tetris_ai_records} , compares representative methods across key metrics: average removed lines, training samples, feature sets, and piece generators. Early approaches like BCTS \cite{thiery2009building}
based on noise cross entropy \cite{szita2006learning}, achieved 3,000 average scores but required \(6.5 \times 10^7\) training samples—an order of magnitude more than later methods such as CBMPI \cite{gabillon2013approximate, scherrer2015approximate} (4,300 avg. with \(8 \times 10^6\) samples) and dSiLU-TD($\lambda$) \cite{elfwing2018sigmoid} (4,900 avg., the highest in the table, with 200,000 samples). Even STEW \cite{lichtenberg2019regularization} (4,800 avg.) relied on a 7-Bag generator (a simplified deterministic setting) to avoid overfitting, sacrificing generalization to classic Tetris’s true stochasticity (i.i.d. piece drawing). While Table \ref{tab:tetris_ai_records} confirms Tetris’s ability to distinguish method performance across sample efficiency and feature robustness, it also exposes a critical gap: high performance often demands massive training resources, while existing open-source implementations (e.g., OpenAI Gym-Tetris \cite{gym-tetris}) suffer from severe runtime inefficiency—taking 12.92 seconds for 10,000 samples vs. 0.24 seconds for our optimized design—hindering large-scale RL training.

Beyond the `performance-efficiency tradeoff' highlighted by Table \ref{tab:tetris_ai_records}, existing Tetris research faces unresolved limitations that restrict its utility for scalable RL.
\begin{itemize}
	\item \textit{Engine inefficiency} plagues grid-based implementations: traditional representations fail to leverage bitwise operations for fast collision detection, line clearing, and state evaluation—core processes that our bitboard design accelerates.
	\item \textit{Policy optimization is constrained}: state-of-the-art methods either depend on complex hand-crafted features (e.g., Bertsekas features \cite{bertsekas1996neuro} in dSiLU-TD($\lambda$) \cite{elfwing2018sigmoid}) or trajectory-based training paradigms that waste resources on low-quality early-stage samples.
\end{itemize}
These issues collectively limit Tetris’s potential to drive progress in sequential decision-making research.

To address these gaps, this paper proposes a high-efficiency Tetris AI framework, with core contributions aligned to the key innovations of the study:
\begin{itemize}
	\item \textit{Bitboard-Based Tetris Implementation}: Redesigning the game board (10 columns as 32-bit integers) and tetrominoes using bitwise operations, achieving a 53-fold speedup over OpenAI Gym-Tetris. This acceleration enables efficient simulation for large-scale RL training by optimizing collision detection, line clearing, and Dellacherie-Thierry (DT) feature calculation.
	\item \textit{Afterstate-Evaluating Actor + Buffer-Optimized PPO}: The afterstate-based Actor simplifies state value estimation by leveraging Tetris’s afterstate property (board configuration post-action but pre-next-piece), outperforming traditional action-value networks. The buffer-optimized PPO reduces training steps to 61,440 (1/1058 of BCTS, 1/3 of dSiLU-TD($\lambda$)) while maintaining a competitive average score of 3,829 on $10\times10$ grids.
	\item \textit{OpenAI Gym-Compliant Python-Java Interface}: Built via Jpype \cite{nelson2020jpype}, this interface bridges Java’s bitwise performance with Python’s RL framework compatibility (e.g., PyTorch, TensorFlow), enabling rapid algorithm prototyping.
\end{itemize}

The rest of this paper is organized as follows: Section 2 covers Tetris background (game mechanics, MDP formalization, DT features). Section 3 details the bitboard implementation and accelerated operations. Section 4 presents afterstate Actor and the buffer-based PPO framework. Section 5 validates performance on $10\times10/10\times20$ grids, comparing to state-of-the-art methods. Section 6 gives the conclusion.

\section{Background}
In this background section, we provide foundational knowledge on the Tetris game and its application in artificial intelligence, emphasizing reinforcement learning frameworks. We first describe the core mechanics of Tetris, including piece manipulation, line clearing, and the critical concept of afterstates, which facilitate efficient state evaluation. Subsequently, we formalize Tetris as a Markov Decision Process (MDP), detailing state and action representations through feature engineering like Dellacherie-Thierry (DT) features. Finally, we discuss policy formulations in Tetris AI, contrasting deterministic greedy approaches with stochastic methods such as Proximal Policy Optimization (PPO) and REINFORCE, setting the stage for our proposed enhancements in environment design and training efficiency.

\subsection{The Tetris Game}
\begin{figure}[!htb]
	\centering
	\begin{tikzpicture}[scale=0.4]
		\foreach \y in {0} {
			\foreach \x in {0,...,9} {
				\fill[blue!60] (\x,\y) rectangle (\x+1,\y+1);
			}
		}
		
		\fill[blue!60] (2,1) rectangle (3,2);
		\fill[blue!60] (2,2) rectangle (3,3);
		\fill[blue!60] (1,2) rectangle (2,3);
		\fill[blue!60] (5,1) rectangle (6,2);
		\fill[blue!60] (7,1) rectangle (8,2);
		\fill[blue!60] (9,1) rectangle (10,2);
		
		\foreach \y in {0,...,20} {
			\draw[thin, black] (0,\y) -- (10,\y);
		}
		
		\foreach \x in {0,...,10} {
			\draw[thin, black] (\x,0) -- (\x,20);
		}
		
		\fill[yellow] (13,17) rectangle (14,18);
		\fill[yellow] (14,17) rectangle (15,18);
		\fill[yellow] (13,18) rectangle (14,19);
		\fill[yellow] (14,18) rectangle (15,19);
		
		\fill[green] (13,15) rectangle (14,16);
		\fill[green] (14,15) rectangle (15,16);
		\fill[green] (15,15) rectangle (16,16);
		\fill[green] (16,15) rectangle (17,16);
		
		\fill[cyan] (14,12) rectangle (15,13);
		\fill[cyan] (15,12) rectangle (16,13);
		\fill[cyan] (13,13) rectangle (14,14);
		\fill[cyan] (14,13) rectangle (15,14);
		
		\fill[red] (13,9) rectangle (14,10);
		\fill[red] (14,9) rectangle (15,10);
		\fill[red] (14,10) rectangle (15,11);
		\fill[red] (15,10) rectangle (16,11);
		
		\fill[orange] (13,6) rectangle (14,7);
		\fill[orange] (14,6) rectangle (15,7);
		\fill[orange] (15,6) rectangle (16,7);
		\fill[orange] (15,7) rectangle (16,8);
		
		\fill[blue] (13,3) rectangle (14,4);
		\fill[blue] (13,4) rectangle (14,5);
		\fill[blue] (14,4) rectangle (15,5);
		\fill[blue] (15,4) rectangle (16,5);
		
		\fill[gray] (14,0) rectangle (15,1);
		\fill[gray] (13,1) rectangle (14,2);
		\fill[gray] (14,1) rectangle (15,2);
		\fill[gray] (15,1) rectangle (16,2);
		
	\end{tikzpicture}
	\caption{Tetris Game Board with All Pieces}
	\label{Fig:tetris_board}
\end{figure}
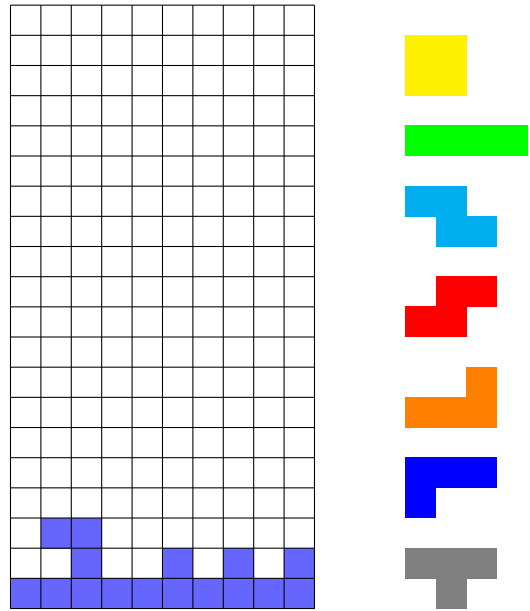

The Tetris game, designed by Alexey Pajitnov and first released in 1984, involves manipulating seven distinct pieces (I, O, T, L, J, S, Z) through translation and rotation to form complete horizontal lines at the bottom of a grid (typically $10 \times 20$, Fig \ref{Fig:tetris_board}). Completed lines are cleared, and blocks above descend, while uncleared lines accumulate until they exceed the grid’s upper boundary, resulting in a game over. The core objective is to maximize the number of cleared lines.

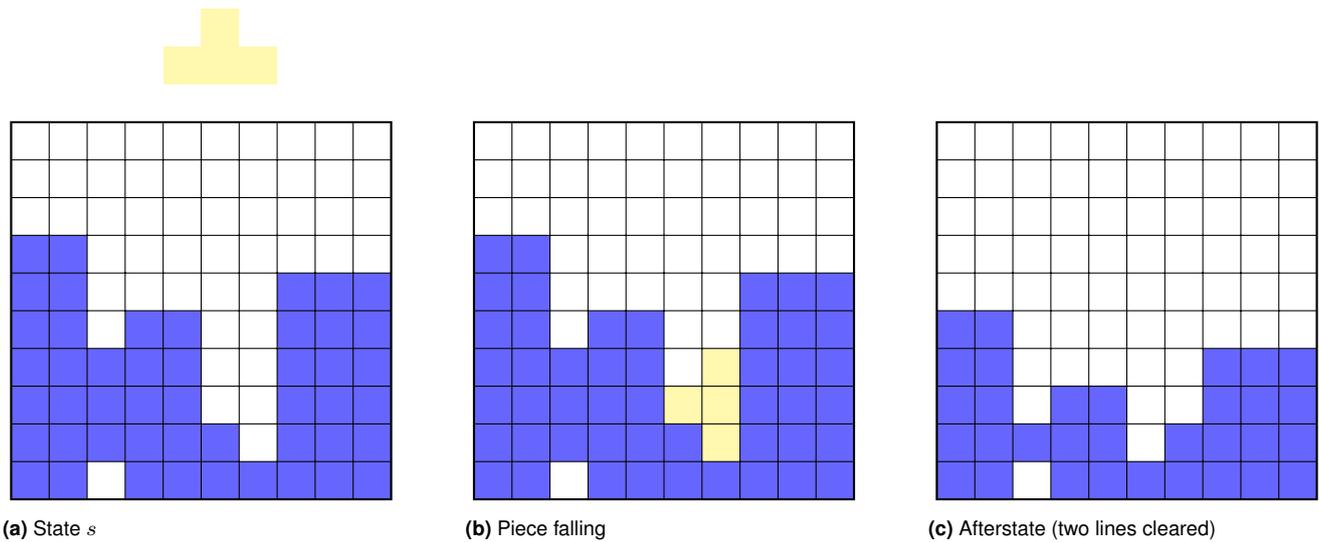
\begin{figure*}[!htb]
	\centering
	\begin{subfigure}[b]{0.3\textwidth}
		\centering
		\begin{tikzpicture}[scale=0.5]
			
			\fill[yellow!40] (4,11) rectangle (5,12);
			\fill[yellow!40] (5,11) rectangle (6,12);
			\fill[yellow!40] (6,11) rectangle (7,12);
			\fill[yellow!40] (5,12) rectangle (6,13);

			\foreach \x/\y in {0/0,0/1,0/2,0/3,0/4,0/5,0/6,
				1/0,1/1,1/2,1/3,1/4,1/5,1/6,
				2/1,2/2,2/3,
				3/0,3/1,3/2,3/3,3/4,
				4/0,4/1,4/2,4/3,4/4,
				5/0,5/1,
				6/0,
				7/0,7/1,7/2,7/3,7/4,7/5,
				8/0,8/1,8/2,8/3,8/4,8/5,
				9/0,9/1,9/2,9/3,9/4,9/5} {
				\fill[blue!60] (\x,\y) rectangle ++(1,1);
			}

			\draw[step=1, thin, black] (0,0) grid (10,10);
			\draw[thick] (0,0) rectangle (10,10);
		\end{tikzpicture}
		\caption{State $s$}
	\end{subfigure}
	\hfill
	\begin{subfigure}[b]{0.3\textwidth}
		\centering
		\begin{tikzpicture}[scale=0.5]
			
			\fill[yellow!40] (5,2) rectangle (6,3);
			\fill[yellow!40] (6,1) rectangle (7,2);
			\fill[yellow!40] (6,2) rectangle (7,3);
			\fill[yellow!40] (6,3) rectangle (7,4);

			\foreach \x/\y in {0/0,0/1,0/2,0/3,0/4,0/5,0/6,
				1/0,1/1,1/2,1/3,1/4,1/5,1/6,
				2/1,2/2,2/3,
				3/0,3/1,3/2,3/3,3/4,
				4/0,4/1,4/2,4/3,4/4,
				5/0,5/1,
				6/0,
				7/0,7/1,7/2,7/3,7/4,7/5,
				8/0,8/1,8/2,8/3,8/4,8/5,
				9/0,9/1,9/2,9/3,9/4,9/5} {
				\fill[blue!60] (\x,\y) rectangle ++(1,1);
			}
			
			\draw[step=1, thin, black] (0,0) grid (10,10);
			\draw[thick] (0,0) rectangle (10,10);
		\end{tikzpicture}
		\caption{Piece falling}
	\end{subfigure}
	\hfill
	\begin{subfigure}[b]{0.3\textwidth}
		\centering
		\begin{tikzpicture}[scale=0.5]
			
			\foreach \x/\y in {0/0,0/1,0/2,0/3,0/4,
				1/0,1/1,1/2,1/3,1/4,
				2/1,
				3/0,3/1,3/2,
				4/0,4/1,4/2,
				5/0,
				6/0,6/1,
				7/0,7/1,7/2,7/3,
				8/0,8/1,8/2,8/3,
				9/0,9/1,9/2,9/3} {
				\fill[blue!60] (\x,\y) rectangle ++(1,1);
			}
			
			\draw[step=1, thin, black] (0,0) grid (10,10);
			\draw[thick] (0,0) rectangle (10,10);
		\end{tikzpicture}
		\caption{Afterstate (two lines cleared)}
		\label{Fig:afterstate}
	\end{subfigure}
	\caption{llustration of the Afterstate in Tetris Game}
	\label{Fig:bitboard_code_case}
\end{figure*}

A critical concept in Tetris is the \textit{afterstate}. Given a state $s$ and an action $a$, the state where the next random piece has not yet been generated after executing action $a$ is called an afterstate. The afterstate reflects the direct result of the current action. As shown in Fig \ref{Fig:bitboard_code_case}, an afterstate (i.e., Fig \ref{Fig:afterstate}) is added before the transition from state $s$ to $s'$. The afterstate combined with the next randomly generated piece constitutes the next state $s'$.This afterstate directly determines the available space for subsequent pieces, making it a key factor in long-term strategy. This creates a perfect scenario for studying RL applications in complex decision-making problems. By designing efficient reward functions and state representations, researchers can explore optimization methods for RL algorithms in high-dimensional state spaces.

\subsection{Tetris AI}
\paragraph{\textbf{MDP Formalization}}
The game of Tetris can be described using the MDP framework to model the interaction between the agent and the game environment. MDP is a fundamental framework in reinforcement learning used to describe the interaction between an agent and its environment. In the game of Tetris, the MDP can be formally represented by a four-tuple $(S, A, R, P)$, which includes the following key elements:

\begin{itemize}
	\item[(i)] \textbf{State Space ($S$)}: Represents the current game situation, including the grid configuration of placed blocks, the type, rotation, and position of the current falling piece, and the next piece in the preview. This defines the complete information set upon which the agent bases its decisions.
	\item[(ii)] \textbf{Action Space ($A$)}: Encompasses all valid operations on the current piece, defined as combinations of rotation states ($0$ to $3$ rotations) and horizontal positions (within the grid boundaries). Each action $(rotation, translation)$ corresponds to a unique placement for the piece.
	\item[(iii)] \textbf{State Transition Probability ($P$)}: Describes the dynamics of the environment. It is a function $P: S \times A \times S \to [0, 1]$. $P(s' | s, a)$ denotes the probability of transitioning to state $s'$ after taking action $a$ in state $s$. In Tetris, transitions are partially deterministic (grid update based on action $a$) and partially stochastic (the next piece is generated randomly with equal probability from the 7 types).
	\item[(iv)] \textbf{Reward Function ($R$)}: Quantifies the immediate feedback for the agent. It is a function $R: S \times A \times S \to \mathbb{R}$. $R(s, a, s')$ represents the scalar reward received after taking action $a$ in state $s$, resulting in state $s'$. Its value is typically determined by the number of lines cleared ($l \in \{0,1,2,3,4\}$) as a direct result of the action.
\end{itemize}

This MDP formulation enables RL agents to learn optimal placement strategies by maximizing cumulative long-term rewards.
In this MDP framework, the agent’s goal is to learn an optimal policy $\pi(a|s)$—a mapping from states to actions—that maximizes the expected cumulative reward. RL algorithms for policy learning are broadly divided into \textit{value-based} (e.g., Q-Learning, which learns action values to derive policies indirectly) and \textit{policy-based} (e.g., PPO, which optimizes the policy directly via gradient ascent), with the latter being more suitable for Tetris’s discrete action space and long-episode training.

\paragraph{\textbf{State and Action Representation}}
Raw state representations (e.g., direct grid arrays) are impractical for Tetris due to their high dimensionality ($10 \times 20 = 200$ binary features). Instead, \textit{feature engineering} simplifies the state by extracting meaningful abstract features. Among the most effective is the DT feature set, which captures critical aspects of board structure and placement quality.

The DT features, building on Dellacherie-Thiery’s 6 original features with 3 extensions \cite{fahey2003tetris, thiery2009improvements}, are defined for a grid of width $w$ and height $h$ (cells $b(x,y) \in \{0,1\}$, where $1$ denotes a filled cell) as follows \cite{amundsen2014comparison}:

\begin{itemize}
	\item[(i)] \textbf{Landing height}: $t_y + (t_h-1)/2$ (vertical distance from the piece’s centroid to the grid bottom, where $t_y$ is the lowest row and $t_h$ is the piece height).
	
	\item[(ii)] \textbf{Eroded piece cells}: $n_r \times n_c$ (contribution to line clears, with $n_r$ = cleared lines and $n_c$ = piece cells in cleared lines).
	
	\item[(iii)] \textbf{Row transitions}: Count of filled/empty transitions in rows (including edges treated as filled, computed via XOR).
	
	\item[(iv)] \textbf{Column transitions}: Vertical counterpart to row transitions, with similar edge handling.
	
	\item[(v)] \textbf{Holes}: Empty cells beneath filled cells in columns, determined using column height $ch(x)$ (maximum filled row in column $x$).
	
	\item[(vi)] \textbf{Board well}: Sum of depths of vertical wells (columns with filled neighbors, boundaries treated as filled).
	
	\item[(vii)] \textbf{Hole depth}: Total distance from holes to the nearest filled cell above.
	
	\item[(viii)] \textbf{Rows with holes}: Count of rows containing at least one hole.
	
	\item[(ix)] \textbf{Pattern diversity}: Number of distinct height differences ($-2$ to $+2$) between adjacent columns.
\end{itemize}

These features are combined linearly to estimate state value: $V(s) = \sum_{i=1}^9 \Phi_i \theta_i$, where $\Phi_i$ are the DT features and $\theta_i$ are learned weights. Two representative weight sets (DT-10 and DT-20) are shown in Table \ref{table_dt1020}.

\begin{table}[htb]
	\centering
	\caption{DT-10 and DT-20 weights \cite{gabillon2013approximate, scherrer2015approximate}.}
	\label{table_dt1020}
		\begin{tabular}{@{}lcc@{}}
			\toprule
			\textbf{Feature} & \textbf{DT-10 weight} & \textbf{DT-20 weight} \\ \midrule
			Landing height & -2.18 & -2.68 \\
			Eroded piece cells & 2.42 & 1.38 \\
			Row transitions & -2.17 & -2.41 \\
			Column transitions & -3.31 & -6.32 \\
			Holes & 0.95 & 2.03 \\
			Board well & -2.22 & -2.71 \\
			Hole depth & -0.81 & -0.43 \\
			Rows with holes & -9.65 & -9.48 \\
			Diversity & 1.27 & 0.89 \\ \bottomrule
	\end{tabular}
\end{table}

For action representation, each piece’s valid placements (rotation + horizontal position) are mapped to discrete indices. For example, if the maximum number of valid actions across all pieces is $N_{\text{max}}$, actions are encoded as $A = \{0, 1, ..., N_{\text{max}} - 1\}$.

\paragraph{\textbf{Policy in Tetris AI}}
The policy function $\pi(a|s)$ defines the probability of selecting action $a$ in state $s$. Policies in Tetris AI can be broadly categorized into deterministic and stochastic types, each tailored to the game's discrete action space and afterstate properties. Common in Tetris RL, deterministic greedy policies select the action that maximizes expected returns:
\begin{equation}
	\pi(a|s) = \arg\max_{a \in A} Q(s,a)
\end{equation}
where $Q(s,a)$ is the action-value function, representing the expected cumulative reward of taking action $a$ in state $s$ and following $\pi$ thereafter. For Dellacherie-Thierry (DT) features, $Q(s,a)$ can be approximated using the state value $V(s')$ of the resulting post-state $s'$:
\begin{equation}
	Q(s,a) \approx R(s,a,s') + \gamma V(s')
\end{equation}
where $R(s,a,s')$ is the immediate reward (lines cleared), $\gamma \in [0,1)$ is the discount factor, and $V(s') = \sum_{i=1}^9 \Phi_i(s') \theta_i$. For stochastic policies, adopted in policy-based algorithms like Proximal Policy Optimization (PPO), the policy is parameterized by a set of parameters $\theta$ (e.g., weights of a neural network) and outputs the probability of selecting action $a$ in state $s$:
\begin{equation}
	\pi_\theta(a|s) = P(a | s; \theta)
\end{equation}
\begin{algorithm}[!htp]
	\floatname{algorithm}{Algorithm}
	\renewcommand{\thealgorithm}{2.1}
	\caption{REINFORCE Algorithm}
	\begin{algorithmic}[1]
		\State Initialize Actor and Critic networks
		\For{each episode}
		\State Compute $\tau$, $done = false$, $s = \text{env.reset()}$
		\While{not $done$}
		\State Compute $f(s,a)_{\text{all}}$ and $mask$ based on state $s$
		\State Actor selects action $a$ using $f(s,a)_{\text{all}}$ and $mask$
		\State Execute $\text{env.step}(a)$, receive next state $s'$ and reward $r$
		\EndWhile
		\State Compute return: $G_t = \sum_{t'=t}^{T} \gamma^{t'-t} r_{t'}$
		\State Update Actor: $\theta \gets \theta + \alpha \sum_{t=0}^{T} G_t \nabla_{\theta} \log \pi_{\theta}(a_t | s_t)$
		\EndFor
	\end{algorithmic}
	\label{Alg:REINFORCE}
\end{algorithm}
In the context of Tetris, typical actor network designs for PPO leverage value function evaluations. The standard approach employs an action value-based actor, where the network evaluates the action-value function for each available action, yielding a Boltzmann-style policy:
\begin{equation}
	\pi(a|s) = \frac{\exp(Q(s, a))}{\sum_b \exp(Q(s, b))}
\end{equation}
This design requires action-dependent inputs (e.g., one-hot encodings of the current tetromino type and action index) in addition to state features. However, a more efficient alternative leverages the afterstate property of Tetris. Since the afterstate abstracts the board configuration post-action but pre-next-piece, it simplifies computation by inherently accounting for expectations over future states: The state-value function of the afterstate represents the expectation over possible next states:
\begin{equation}\label{Eq:as_exp}
	V(as) = \sum_{s' \in \mathcal{S}} V(s') P(s'|as)
\end{equation}
This leads to the afterstate-based actor, where actions are selected according to:
\begin{equation}\label{Eq:Actor_vas}
	\pi(a|s) = \frac{\exp(V(f(s, a)))}{\sum_b \exp(V(f(s, b)))}
\end{equation}
Here, $f(s,a)$ represents the feature representation of the afterstate reached by executing action $a$ in state $s$. This approach maintains a simpler network architecture while incorporating uncertainty over future pieces. PPO~\cite{schulman2017PPO} is a policy gradient method that updates policy network parameters by maximizing the expected cumulative reward. Its core innovation is the use of a clipped surrogate objective to constrain update magnitudes, ensuring similarity between new and old policies for enhanced training stability and efficiency. This addresses high variance in policy gradients, a common issue in methods like REINFORCE (detailed below). Compared to Trust Region Policy Optimization (TRPO)~\cite{schulman2015TRPO}, PPO reduces computational complexity by approximating the KL-divergence constraint with clipping, making it more scalable for games like Tetris. In this work, we adopt a trajectory-based PPO algorithm as the baseline to train the Tetris agent: for each update, we sample a complete game trajectory to refine the policy network. The specific performance of this trajectory-based PPO baseline (\textbf{Algorithm~\ref{Alg:PPO_trajectory}}) will be elaborated on in the Experiments section. For comparison, we also consider REINFORCE(\textbf{Algorithm~\ref{Alg:REINFORCE}}) as a simpler policy gradient baseline. REINFORCE updates the policy using Monte Carlo estimates of returns, but suffers from high variance without variance reduction techniques like baselines or advantages:

\begin{algorithm}[htp]
	\floatname{algorithm}{Algorithm}
	\renewcommand{\thealgorithm}{2.2}
	\caption{Trajectory-based PPO Algorithm}
	\footnotesize
	\begin{algorithmic}[1]
		\State Initialize Actor and Critic networks
		\For{each episode}
		\State Set $t = 0$, $done = false$, $s = \text{env.reset()}$
		\While{not $done$}
		\State Compute $f(s,a)_{\text{all}}$ and $mask$ based on state $s$
		\State Actor selects action $a$ using $f(s,a)_{\text{all}}$ and $mask$
		\State Execute $\text{env.step}(a)$, receive next state $s'$ and reward $r$
		\EndWhile
		\State Compute TD error:
		\State \quad $\delta_t = r + \gamma \cdot \text{Critic}(f(s_t, a_t)) - \text{Critic}(f(s_t))$
		\State Compute Generalized Advantage Estimation (GAE):
		\State \quad $A_t = \sum_{l=0}^{\infty} (\gamma \lambda)^l \delta_{t+l}$
		\For{each training epoch $k$}
		\State Update Actor:
		\State \quad $\theta \gets \theta + \alpha_\theta \nabla_\theta L_{\text{a}}$
		\State \quad where $L_{\text{a}} = \min\left[ \frac{\pi_{\theta}(a|s)}{\pi_{\theta_k}(a|s)} A_t, \right.$
		\State \quad \quad $\left. \text{clip}\left( \frac{\pi_{\theta}(a|s)}{\pi_{\theta_k}(a|s)}, 1-\epsilon, 1+\epsilon \right) A_t \right]$
		\State Update Critic:
		\State \quad $w \gets w + \alpha_w \sum_{t} \delta_t \nabla_w V_w(s_t)$
		\EndFor
		\EndFor
	\end{algorithmic}
	\normalsize
	\label{Alg:PPO_trajectory}
\end{algorithm}

\section{Bitboard of the Tetris Game}

In this section, we first present the design of Tetris based on bitboard, including the representation of the game board and tetrominoes, as well as the acceleration of various game operations through bitwise operations. Subsequently, we introduce the formal definition of DT features and provide their implementation using bitwise operations. Considering language execution efficiency and type constraints, Java was ultimately selected to implement the bitboard-based Tetris, while also providing a Python interface compliant with the OpenAI Gym standard \cite{gym-tetris}.

\subsection{Bitboard Representation for the Tetris Game}
Inspired by the Bitboard concept, the Tetris game board can be redesigned. A 32-bit integer is used to represent the state of a single column on the board. For a column, if there is a block at a certain position from bottom to top, the corresponding bit in the integer from the least significant bit to the most significant bit is set to 1. As shown in Figure \ref{Fig:tetris_board}, in the second column from the left, the first and third rows from the bottom contain blocks. Therefore, the second column can be represented in binary as 0B101, which is 5 in decimal, indicating that there are blocks in the first and third rows of that column. Consequently, a single column of the board can be represented by a 32-bit integer, and the entire board can be represented by a 10-dimensional array of int-type integers. This design approach is applicable to both $10\times10$ and $10\times20$ game boards.

There are seven types of Tetris pieces: O, I, S, Z, L, J, and T. Following the design approach for the Tetris game board, the pieces also need to be redesigned based on Bitboard. The definition of the Piece class is shown in Code \ref{code:Sq_def}.

\begin{figure}[!htb]
	\centering
	\begin{minipage}{0.9\linewidth}
		\begin{lstlisting}[
			language=Java, 
			caption={Piece Definition Class}, 
			label={code:Sq_def}
			]
public static class Piece {
	public int[] piece; 
	public int width;   
	public int height;  
	
	public Piece(int[] p, int w, int h) {
		piece = p;
		width = w;
		height = h;
	}
}

public static class Sq {
	public Piece[] pieces;
	public int rotationNum;  
	public int actionSize;   
	public int maxHeight;    
	
	public Sq(Piece[] p, int s, int size, int height) {
		pieces = p;
		rotationNum = s;
		actionSize = size;
		maxHeight = height;
	}
}
		\end{lstlisting}
	\end{minipage}
\end{figure}

\begin{figure}[htb]
	\centering
	\begin{tikzpicture}[scale=0.6]
		\draw[fill=yellow] (0,0) rectangle (1,1);
		\draw[fill=yellow] (0,1) rectangle (1,2);
		\draw[fill=yellow] (1,0) rectangle (2,1);
		\draw[fill=yellow] (1,1) rectangle (2,2);
	\end{tikzpicture}
	\caption{The O-Piece and Its Rotation(s)}
	\label{Fig:O_Piece}
\end{figure}
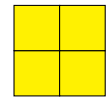

The Piece class defines a particular shape of a piece after rotation. The Piece class has three properties: an int array representing the piece, width indicating the width of the piece in that shape, and height indicating the height of the piece in that shape. Since a piece can have different shapes under different rotations, the Sq class is used for encapsulation, packaging all different shapes of the same piece into a single Sq object. The Piece array is used to store the different shapes of the piece after rotation, with the array length determined by the number of shapes the piece can take. rotationNum represents the number of rotations the piece can undergo, actionSize represents the maximum number of actions available for the piece, and maxHeight represents the maximum height of the piece in different shapes. In addition to the basic form of the piece, the width and height of the piece in that shape are also stored for convenience in subsequent DT feature calculations.
Take the O piece as an example. As shown in Figure \ref{Fig:O_Piece}, the first and second columns are both 0B11, so the O piece can be represented as {0B11, 0B11}, i.e., {3, 3}, and stored in an int-type array.

The code implementation for the O-piece is shown in Code \ref{lst:O-piece}. It is evident that regardless of rotation, the O-piece has only one shape, i.e., {3, 3}. The height and width for this shape are both 2, so only one piece needs to be defined. Since the O-piece has a fixed shape under different rotations, when considering the maximum number of available actions, only horizontal movement needs to be taken into account. On a 10-column grid, it can have up to nine different placement positions.

\begin{figure}[!htb]
	\centering
	\begin{minipage}{0.9\linewidth} 
		\begin{lstlisting}[
			language=Java, 
			caption={Bitboard Representation of O-Piece}, 
			label={lst:O-piece},
			]
int[] O_piece1 = {3, 3}; // O-piece
Piece[] piece1 = new Piece[1];
piece1[0] = new Piece(O_piece1, 2, 2);
squares[0] = new Sq(piece1, 1, 9, 2);
			
		\end{lstlisting}
	\end{minipage}
\end{figure}

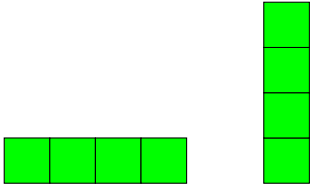
\begin{figure}[ht]
	\centering
	\begin{minipage}[t]{0.2\linewidth}
		\centering
		\begin{tikzpicture}[scale=0.6]
			\foreach \x in {0,1,2,3} {
				\draw[fill=green] (\x,0) rectangle (\x + 1,1);
			}
		\end{tikzpicture}
	\end{minipage}
	\hspace{1cm}
	\begin{minipage}[t]{0.2\linewidth}
		\centering
		\begin{tikzpicture}[scale=0.6]
			\foreach \y in {0,1,2,3} {
				\draw[fill=green] (0,\y) rectangle (1,\y + 1);
			}
		\end{tikzpicture}
	\end{minipage}
	\caption{The I-Piece and Its Rotations}
	\label{Fig:I_Piece}
\end{figure}

Unlike the O-piece, the I-piece has two shapes: horizontal and vertical, as shown in Figure \ref{Fig:I_Piece} for their Bitboard representations.

The code implementation of the I-piece is shown in Listing \ref{lst:I-piece}. \text{I\_piece1} represents the vertical orientation of the I-piece, which can be denoted by \{0B1111\}, or equivalently \{15\}. \text{I\_piece2} represents the horizontal orientation of the I-piece, which can be denoted by \{0B1, 0B1, 0B1, 0B1\}, or equivalently \{1, 1, 1, 1\}. In this case, the length of the Piece class array is 2, storing the bitboard representations and the corresponding heights and widths for both the horizontal and vertical orientations. The maximum number of selectable actions is 17 in total, comprising 7 actions in the horizontal orientation and 10 actions in the vertical orientation.

\begin{figure}[!htb]
	\centering
	\begin{minipage}{0.9\linewidth}
		\begin{lstlisting}[
			language=Java, 
			caption={Bitboard Representation of I-Piece}, 
			label={lst:I-piece},
			]
int[] I_piece1 = {15};     // I-piece
int[] I_piece2 = {1, 1, 1, 1};
Piece[] piece2 = new Piece[2];
piece2[0] = new Piece(I_piece1, 1, 4);
piece2[1] = new Piece(I_piece2, 4, 1);
squares[1] = new Sq(piece2, 2, 17, 4);
		\end{lstlisting}
	\end{minipage}
\end{figure}

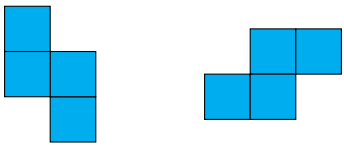
\begin{figure}[ht]
	\centering
	\begin{minipage}{0.2\linewidth}
		\centering
		\begin{tikzpicture}[scale=0.6]
			\draw[fill=cyan] (1,0) rectangle (2,1);
			\draw[fill=cyan] (1,1) rectangle (2,2);
			\draw[fill=cyan] (0,1) rectangle (1,2);
			\draw[fill=cyan] (0,2) rectangle (1,3);
		\end{tikzpicture}
	\end{minipage}
	\hspace{1cm}
	\begin{minipage}{0.2\linewidth}
		\centering
		\begin{tikzpicture}[scale=0.6]
			\draw[fill=cyan] (0,0) rectangle (1,1);
			\draw[fill=cyan] (1,1) rectangle (2,2);
			\draw[fill=cyan] (1,0) rectangle (2,1);
			\draw[fill=cyan] (2,1) rectangle (3,2);
		\end{tikzpicture}
	\end{minipage}
	\caption{The S-Piece and Its Rotations}
	\label{Fig:S_Piece}
\end{figure}

The S-piece has two shapes, as shown in Figure \ref{Fig:S_Piece} for their Bitboard representations.

The code implementation of the S-piece is shown in Listing \ref{lst:S-piece}. \text{S\_piece1} represents the vertical orientation on the left in Figure \ref{Fig:S_Piece}, which can be denoted by \{0B110, 0B011\}, or equivalently \{6, 3\}. \text{S\_piece2} represents the horizontal orientation on the right in Figure \ref{Fig:S_Piece}, which can be denoted by \{0B01, 0B11, 0B10\}, or equivalently \{1, 3, 2\}. In this case, the length of the Piece class array is 2. The maximum number of selectable actions is 17 in total, comprising 9 actions in the vertical orientation and 8 actions in the horizontal orientation.

\begin{figure}[!htb]
	\centering
	\begin{minipage}{0.9\linewidth}
		\begin{lstlisting}[
			language=Java, 
			caption={Bitboard Representation of S-Piece}, 
			label={lst:S-piece},
			]
int[] S_piece1 = {6, 3};          // S-piece
int[] S_piece2 = {1, 3, 2};
Piece[] piece3 = new Piece[2];
piece3[0] = new Piece(S_piece1, 2, 3);
piece3[1] = new Piece(S_piece2, 3, 2);
squares[2] = new Sq(piece3, 2, 17, 3);
		\end{lstlisting}
	\end{minipage}
\end{figure}

\begin{figure}[!ht]
	\centering
	\begin{minipage}[t]{0.2\linewidth}
		\centering
		\begin{tikzpicture}[scale=0.6]
			\draw[fill=red] (0,0) rectangle (1,1);
			\draw[fill=red] (0,1) rectangle (1,2);
			\draw[fill=red] (1,1) rectangle (2,2);
			\draw[fill=red] (1,2) rectangle (2,3);
		\end{tikzpicture}
	\end{minipage}
	\hspace{1cm}
	\begin{minipage}[t]{0.2\linewidth}
		\centering
		\begin{tikzpicture}[scale=0.6]
			\draw[fill=red] (0,1) rectangle (1,2);
			\draw[fill=red] (1,1) rectangle (2,2);
			\draw[fill=red] (1,0) rectangle (2,1);
			\draw[fill=red] (2,0) rectangle (3,1);
		\end{tikzpicture}
	\end{minipage}
	\caption{The Z-Piece and Its Rotations}
	\label{Fig:Z_Piece}
\end{figure}
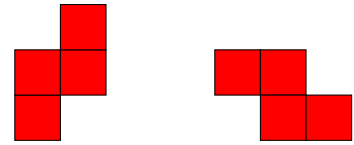

The Z-piece has two shapes, as shown in Figure \ref{Fig:Z_Piece} for their Bitboard representations.

The code implementation for the Z-piece is shown in Code \ref{lst:Z-piece}. \text{Z\_piece1} corresponds to the vertical placement of the Z-piece shown on the left in Figure \ref{Fig:Z_Piece} and can be represented as $\{\text{0B}011, \text{0B}110\}$, i.e., $\{3, 6\}$. \text{Z\_piece2} corresponds to the horizontal placement of the Z-piece shown on the right in Figure \ref{Fig:Z_Piece} and can be represented as $\{\text{0B}10, \text{0B}11, \text{0B}01\}$, i.e., $\{2, 3, 1\}$. At this point, the Piece class array has a length of 2. The maximum number of available actions is 17, consisting of nine for the vertical placement and eight for the horizontal placement.

\begin{figure}[!htb]
	\centering
	\begin{minipage}{0.9\linewidth} 
		\begin{lstlisting}[
			language=Java, 
			caption={Bitboard Representation of Z-Piece}, 
			label={lst:Z-piece},
			]
int[] Z_piece1 = {3, 6};         // Z-piece
int[] Z_piece2 = {2, 3, 1};
Piece[] piece4 = new Piece[2];
piece4[0] = new Piece(Z_piece1, 2, 3);
piece4[1] = new Piece(Z_piece2, 3, 2);
squares[3] = new Sq(piece4, 2, 17, 3);
		\end{lstlisting}
	\end{minipage}
\end{figure}

\begin{figure}[!ht]
	\centering
	\begin{minipage}[t]{0.2\linewidth}
		\centering
		\begin{tikzpicture}[scale=0.6]
			\draw[fill=orange] (0,0) rectangle (1,1);
			\draw[fill=orange] (0,1) rectangle (1,2);
			\draw[fill=orange] (0,2) rectangle (1,3);
			\draw[fill=orange] (1,0) rectangle (2,1);
		\end{tikzpicture}
		\vspace{8pt} \\
		(a)
	\end{minipage}
	\hspace{1cm}
	\begin{minipage}[t]{0.2\linewidth}
		\centering
		\begin{tikzpicture}[scale=0.6]
			\draw[fill=orange] (0,0) rectangle (1,1);
			\draw[fill=orange] (0,1) rectangle (1,2);
			\draw[fill=orange] (1,1) rectangle (2,2);
			\draw[fill=orange] (2,1) rectangle (3,2);
		\end{tikzpicture}
		\vspace{-4pt} \\
		(b)
	\end{minipage}
	\\ \vspace{12pt}
	\begin{minipage}[t]{0.2\linewidth}
		\centering
		\begin{tikzpicture}[scale=0.6]
			\draw[fill=orange] (0,2) rectangle (1,3);
			\draw[fill=orange] (1,0) rectangle (2,1);
			\draw[fill=orange] (1,1) rectangle (2,2);
			\draw[fill=orange] (1,2) rectangle (2,3);
		\end{tikzpicture}
		\vspace{8pt} \\
		(c)
	\end{minipage}
	\hspace{1cm}
	\begin{minipage}[t]{0.2\linewidth}
		\centering
		\begin{tikzpicture}[scale=0.6]
			\draw[fill=orange] (0,0) rectangle (1,1);
			\draw[fill=orange] (1,0) rectangle (2,1);
			\draw[fill=orange] (2,0) rectangle (3,1);
			\draw[fill=orange] (2,1) rectangle (3,2);
		\end{tikzpicture}
		\vspace{-4pt} \\
		(d)
	\end{minipage}
	\caption{The L-Piece and Its Rotations} 
	\label{Fig:L_Piece}
\end{figure}
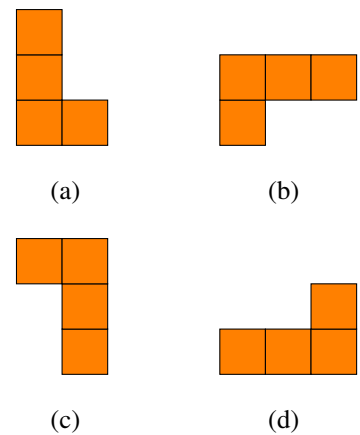

The L-piece has four shapes, as shown in Figure \ref{Fig:L_Piece} for their Bitboard representations.

The code implementation for the L-piece is shown in Code \ref{lst:L-piece}. 
\text{L\_piece1} corresponds to Figure \ref{Fig:L_Piece}(a) and can be represented as 
$\{\text{0B}111, \text{0B}001\}$, i.e., $\{7, 1\}$, with nine available actions. 
\text{L\_piece2} corresponds to Figure \ref{Fig:L_Piece}(b) and can be represented as 
$\{\text{0B}11, \text{0B}10, \text{0B}10\}$, i.e., $\{3, 2, 2\}$, with eight available actions. 
\text{L\_piece3} corresponds to Figure \ref{Fig:L_Piece}(c) and can be represented as 
$\{\text{0B}100, \text{0B}111\}$, i.e., $\{4, 7\}$, with nine available actions. 
\text{L\_piece4} corresponds to Figure \ref{Fig:L_Piece}(d) and can be represented as 
$\{\text{0B}01, \text{0B}01, \text{0B}11\}$, i.e., $\{1, 1, 3\}$, with eight available actions. 
The maximum number of available actions is 34.

\begin{figure}[!htb]
	\centering
	\begin{minipage}{0.9\linewidth} 
		\begin{lstlisting}[
			language=Java, 
			caption={Bitboard Representation of L-Piece}, 
			label={lst:L-piece},
			]
int[] L_piece1 = {7, 1};          // L-piece
int[] L_piece2 = {3, 2, 2};
int[] L_piece3 = {4, 7};
int[] L_piece4 = {1, 1, 3};
Piece[] piece5 = new Piece[4];
piece5[0] = new Piece(L_piece1, 2, 3);
piece5[1] = new Piece(L_piece2, 3, 2);
piece5[2] = new Piece(L_piece3, 2, 3);
piece5[3] = new Piece(L_piece4, 3, 2);
squares[4] = new Sq(piece5, 4, 34, 3);
		\end{lstlisting}
	\end{minipage}
\end{figure}

\begin{figure}[!htb]
	\centering
	\begin{minipage}[t]{0.2\linewidth}
		\centering
		\begin{tikzpicture}[scale=0.6]
			\draw[fill=blue] (0,0) rectangle (1,1);
			\draw[fill=blue] (1,0) rectangle (2,1);
			\draw[fill=blue] (1,1) rectangle (2,2);
			\draw[fill=blue] (1,2) rectangle (2,3);
		\end{tikzpicture}
		\vspace{8pt} \\
		(a)
	\end{minipage}
	\hspace{1cm}
	\begin{minipage}[t]{0.2\linewidth}
		\centering
		\begin{tikzpicture}[scale=0.6]
			\draw[fill=blue] (0,0) rectangle (1,1);
			\draw[fill=blue] (0,1) rectangle (1,2);
			\draw[fill=blue] (1,0) rectangle (2,1);
			\draw[fill=blue] (2,0) rectangle (3,1);
		\end{tikzpicture}
		\vspace{-4pt} \\
		(b)
	\end{minipage}
	\\ \vspace{12pt}
	\begin{minipage}[t]{0.2\linewidth}
		\centering
		\begin{tikzpicture}[scale=0.6]
			\draw[fill=blue] (0,0) rectangle (1,1);
			\draw[fill=blue] (0,1) rectangle (1,2);
			\draw[fill=blue] (0,2) rectangle (1,3);
			\draw[fill=blue] (1,2) rectangle (2,3);
		\end{tikzpicture}
		\vspace{8pt} \\
		(c)
	\end{minipage}
	\hspace{1cm}
	\begin{minipage}[t]{0.2\linewidth}
		\centering
		\begin{tikzpicture}[scale=0.6]
			\draw[fill=blue] (0,1) rectangle (1,2);
			\draw[fill=blue] (1,1) rectangle (2,2);
			\draw[fill=blue] (2,0) rectangle (3,1);
			\draw[fill=blue] (2,1) rectangle (3,2);
		\end{tikzpicture}
		\vspace{-4pt} \\
		(d)
	\end{minipage}
	\caption{The J-Piece and Its Rotations} 
	\label{Fig:J_Piece}
\end{figure}
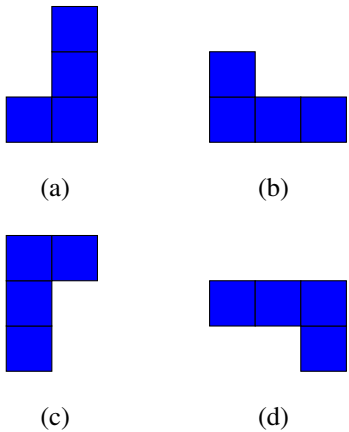

The J-piece has four shapes, as shown in Figure \ref{Fig:J_Piece} for their Bitboard representations.

The code implementation for the J-piece is shown in Code \ref{lst:J-piece}. 
\text{J\_piece1} corresponds to Figure \ref{Fig:J_Piece}(a) and can be represented as 
$\{\text{0B}001, \text{0B}111\}$, i.e., $\{1, 7\}$, with nine available actions. 
\text{J\_piece2} corresponds to Figure \ref{Fig:J_Piece}(b) and can be represented as 
$\{\text{0B}11, \text{0B}01, \text{0B}01\}$, i.e., $\{3, 1, 1\}$, with eight available actions. 
\text{J\_piece3} corresponds to Figure \ref{Fig:J_Piece}(c) and can be represented as 
$\{\text{0B}111, \text{0B}100\}$, i.e., $\{7, 4\}$, with nine available actions. 
\text{J\_piece4} corresponds to Figure \ref{Fig:J_Piece}(d) and can be represented as 
$\{\text{0B}01, \text{0B}01, \text{0B}11\}$, i.e., $\{2, 2, 3\}$, with eight available actions. 
The maximum number of available actions is 34.

\begin{figure}[!htb]
	\centering
	\begin{minipage}{0.9\linewidth} 
		\begin{lstlisting}[
			language=Java, 
			caption={Bitboard Representation of J-Piece}, 
			label={lst:J-piece},
			]
int[] J_piece1 = {1, 7};   // J-piece
int[] J_piece2 = {3, 1, 1};
int[] J_piece3 = {7, 4};
int[] J_piece4 = {2, 2, 3};
Piece[] piece6 = new Piece[4];
piece6[0] = new Piece(J_piece1, 2, 3);
piece6[1] = new Piece(J_piece2, 3, 2);
piece6[2] = new Piece(J_piece3, 2, 3);
piece6[3] = new Piece(J_piece4, 3, 2);
squares[5] = new Sq(piece6, 4, 34, 3);
		\end{lstlisting}
	\end{minipage}
\end{figure}

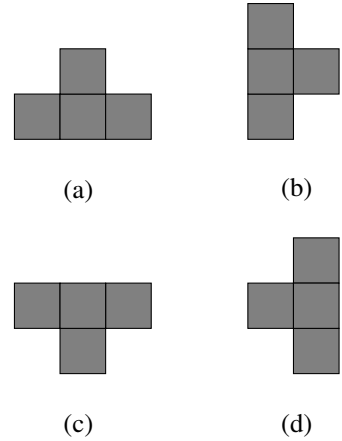
\begin{figure}[!ht]
	\centering
	\begin{minipage}[t]{0.2\linewidth}
		\centering
		\begin{tikzpicture}[scale=0.6]
			\draw[fill=gray] (0,0) rectangle (1,1);
			\draw[fill=gray] (1,0) rectangle (2,1);
			\draw[fill=gray] (1,1) rectangle (2,2);
			\draw[fill=gray] (2,0) rectangle (3,1);
		\end{tikzpicture}
		\vspace{-2pt} \\
		(a)
	\end{minipage}
	\hspace{1cm}
	\begin{minipage}[t]{0.2\linewidth}
		\centering
		\begin{tikzpicture}[scale=0.6]
			\draw[fill=gray] (0,0) rectangle (1,1);
			\draw[fill=gray] (0,1) rectangle (1,2);
			\draw[fill=gray] (0,2) rectangle (1,3);
			\draw[fill=gray] (1,1) rectangle (2,2);
		\end{tikzpicture}
		\vspace{9pt} \\
		(b)
	\end{minipage}
	\\ \vspace{12pt}
	\begin{minipage}[t]{0.2\linewidth}
		\centering
		\begin{tikzpicture}[scale=0.6]
			\draw[fill=gray] (0,1) rectangle (1,2);
			\draw[fill=gray] (1,0) rectangle (2,1);
			\draw[fill=gray] (2,1) rectangle (3,2);
			\draw[fill=gray] (1,1) rectangle (2,2);
		\end{tikzpicture}
		\vspace{-2pt} \\
		(c)
	\end{minipage}
	\hspace{1cm}
	\begin{minipage}[t]{0.2\linewidth}
		\centering
		\begin{tikzpicture}[scale=0.6]
			\draw[fill=gray] (0,1) rectangle (1,2);
			\draw[fill=gray] (1,0) rectangle (2,1);
			\draw[fill=gray] (1,2) rectangle (2,3);
			\draw[fill=gray] (1,1) rectangle (2,2);
		\end{tikzpicture}
		\vspace{10pt} \\
		(d)
	\end{minipage}
	\caption{The T-Piece and Its Rotations} 
	\label{Fig:T_Piece}
\end{figure}

The T-piece has four shapes, as shown in Figure \ref{Fig:T_Piece} for their Bitboard representations.

The code implementation for the T-piece is shown in Code \ref{lst:T-piece}. 
\text{T\_piece1} corresponds to Figure \ref{Fig:T_Piece}(a) and can be represented as 
$\{\text{0B}01, \text{0B}11, \text{0B}01\}$, i.e., $\{1, 3, 1\}$, with eight available actions. 
\text{T\_piece2} corresponds to Figure \ref{Fig:T_Piece}(b) and can be represented as 
$\{\text{0B}111, \text{0B}010\}$, i.e., $\{7, 2\}$, with nine available actions. 
\text{T\_piece3} corresponds to Figure \ref{Fig:T_Piece}(c) and can be represented as 
$\{\text{0B}10, \text{0B}11, \text{0B}10\}$, i.e., $\{2, 3, 2\}$, with eight available actions. 
\text{T\_piece4} corresponds to Figure \ref{Fig:T_Piece}(d) and can be represented as 
$\{\text{0B}010, \text{0B}111\}$, i.e., $\{2, 7\}$, with nine available actions. 
The maximum number of available actions is 34.

\begin{figure}[!htb]
	\centering
	\begin{minipage}{0.9\linewidth} 
		\begin{lstlisting}[
			language=Java, 
			caption={Bitboard Representation of T-Piece}, 
			label={lst:T-piece},
			]
int[] T_piece1 = {1, 3, 1};          // T-piece
int[] T_piece2 = {7, 2};
int[] T_piece3 = {2, 3, 2};
int[] T_piece4 = {2, 7};
Piece[] piece7 = new Piece[4];
piece7[0] = new Piece(T_piece1, 3, 2);
piece7[1] = new Piece(T_piece2, 2, 3);
piece7[2] = new Piece(T_piece3, 3, 2);
piece7[3] = new Piece(T_piece4, 2, 3);
squares[6] = new Sq(piece7, 4, 34, 3);
		\end{lstlisting}
	\end{minipage}
\end{figure}

So far, this paper has presented the Bitboard-based design for the seven types of pieces in the Tetris game. Building upon this representation, the next critical step is to efficiently simulate piece movements and landing detection through bitwise operations.

\paragraph{\textbf{Piece Landing}}
During the Tetris game process, one of the seven types of pieces is randomly generated. When a piece lands, it must make contact with the existing pieces on the board but not overlap with them. A common approach is to simulate the piece starting from the highest column height within the falling range and descending one step at a time. After each descent, it checks whether any cell overlaps with the existing pieces on the board. If an overlap occurs, the previous position is the actual landing spot of the piece.
From the above process, it is evident that the highest column height within the falling range needs to be quickly identified. Below is an introduction to the algorithm for calculating the column height of a single column. In practice, iterating through all columns within the piece's falling range will yield the highest column height within that range. The pseudocode is shown in {\bf{Algorithm \ref{Alg:cal_height}}}, with the input being a column's data of type int (a 4-byte integer with 32 bits), and the output being the column height minus 1.

\begin{algorithm}[!htp]
	\floatname{algorithm}{Algorithm}
	\renewcommand{\thealgorithm}{3.1}
	\caption{Binary Search for Column Height}
	\begin{algorithmic}[1]
		\renewcommand{\algorithmicrequire}{\textbf{Input}:}
		\renewcommand{\algorithmicensure}{\textbf{Output}:}	
		\Require Column data $d$ of the int type
		\State Initialize $h=0$
		\If{($ d\ \& $\ 0xffff0000 $\neq 0 $)}
		\State $d >>> 16$
		\State $h$ += $16$
		\EndIf
		\If{($d\ \& \ $0xff00 $\neq 0$)}
		\State $d >>> 8$
		\State $h$ += $8$
		\EndIf
		\If{($d\ \& \ $0xf0 $\neq 0$)}
		\State $d >>> 4$
		\State $h$ += $4$
		\EndIf
		\If{($d\ \& \ $0x0c $\neq 0$)}
		\State $d >>> 2$
		\State $h$ += $2$
		\EndIf
		\If{($d\ \& \ $0x02 $\neq 0$)}
		\State $h$ += $1$
		\EndIf
		\Ensure $h$
	\end{algorithmic}
	\label{Alg:cal_height}
\end{algorithm}

\begin{figure}[!htb]
	\centering
	\subfloat[Shift the piece up by column height positions]{
		\label{Fig:collision_det:a} 
		\begin{tikzpicture}[scale=0.7][scale=1]
			\fill[blue!60] (0,0) rectangle (1,1);
			\fill[blue!60] (1,0) rectangle (2,1);
			\fill[blue!60] (1,1) rectangle (2,2);
			\fill[blue!60] (2,0) rectangle (3,1);
			\fill[blue!60] (3,0) rectangle (4,1);
			\fill[blue!60] (4,0) rectangle (5,1);
			
			\draw[fill=orange] (1,2) rectangle (2,3);
			\draw[fill=orange] (1,3) rectangle (2,4);
			\draw[fill=orange] (1,4) rectangle (2,5);
			\draw[fill=orange] (0,4) rectangle (1,5);
			\foreach \y in {0,...,5} {
				\draw[thin, black] (0,\y) -- (5,\y);
			}
			\foreach \x in {0,...,5} {
				\draw[thin, black] (\x,0) -- (\x,5);
			}
		\end{tikzpicture}
	}
	\hspace{0.5cm}
	\subfloat[Shift the piece up by column height minus 1 positions]{
		\label{Fig:collision_det:b} 
		\begin{tikzpicture}[scale=0.7][scale=1]
			\fill[blue!60] (0,0) rectangle (1,1);
			\fill[blue!60] (1,0) rectangle (2,1);
			\fill[blue!60] (1,1) rectangle (2,2);
			\fill[blue!60] (2,0) rectangle (3,1);
			\fill[blue!60] (3,0) rectangle (4,1);
			\fill[blue!60] (4,0) rectangle (5,1);
			
			\draw[fill=orange] (1,1) rectangle (2,2);
			\draw[fill=orange] (1,2) rectangle (2,3);
			\draw[fill=orange] (1,3) rectangle (2,4);
			\draw[fill=orange] (0,3) rectangle (1,4);
			\foreach \y in {0,...,5} {
				\draw[thin, black] (0,\y) -- (5,\y);
			}
			\foreach \x in {0,...,5} {
				\draw[thin, black] (\x,0) -- (\x,5);
			}
		\end{tikzpicture}
	}
	\caption{Collision Detection}
	\label{Fig:collision_det}
\end{figure}
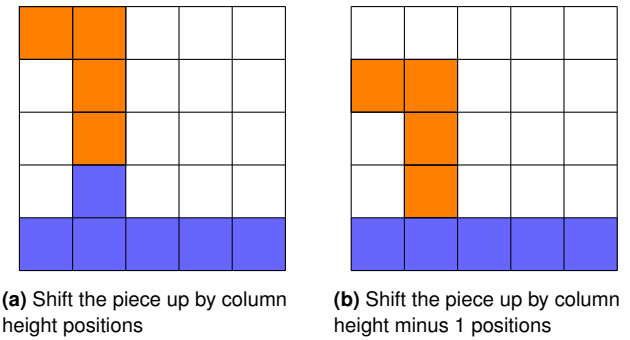

This algorithm outputs the column height minus 1 rather than the original height value to optimize the collision detection process. As shown in Figure \ref{Fig:collision_det}, if the L-piece is placed in columns one and two in the Bitboard form {4, 7}, with column one having a height of 2 and column two a height of 0. Shifting each column's value of the piece left by 2 (equivalent to moving the piece up two positions from the bottom of the board) results in {16, 28}, as shown in Figure \ref{Fig:collision_det:a}. Clearly, no collision detection is needed in this case. As shown in Figure \ref{Fig:collision_det:b}, collision detection should actually begin by moving the piece up by column height minus 1 positions, i.e., the bottommost cell of the falling piece should be less than or equal to the highest column.

Another issue to address is how to quickly perform collision detection. Thanks to the Bitboard design of the game pieces and board, collision detection can be achieved by performing a bitwise AND (\&) operation between the column value of the board and the column value of the piece. If the result is not zero, it indicates a collision. Taking Figure~\ref{Fig:collision_det:b} as an example, the board's first two columns are in state $\{0B00001, 0B00011\}$, and the piece's state is $\{0B01000, 0B01110\}$. After performing the bitwise AND operation, the result is $\{0B00000, 0B00010\}$. Since the result of the bitwise AND operation in the second column is not zero, it can be concluded that an overlap has occurred between the piece and the board.

\paragraph{\textbf{Piece Elimination}}
After a piece lands, it is necessary to check for line clearing operations. The detection process is relatively simple: a variable deleteLine initialized to 0xfffff (i.e., 20 ones in decimal, with the number of ones equal to the column height) is used. By performing a bitwise AND operation between the deleteLine variable and each column of the board, the rows that need to be cleared are identified as the positions where the bits are 1 in the final result. For a board with a column height of 20, the pseudocode for line clearing is shown in {\bf{Algorithm \ref{Alg:delete_line}}}, where the int-type state array stores the states of the ten columns. Through belowTemp, the blocks below the line to be cleared are preserved in below, and through aboveTemp, the blocks above the line to be cleared are preserved in above. By performing a bitwise AND operation between below and (above $>>$1), the column after clearing the line is obtained. This process is repeated until no more lines need to be cleared.

\begin{algorithm}[!htp]
	\floatname{algorithm}{Algorithm}
	\renewcommand{\thealgorithm}{3.2}
	\caption{Line Clearing}
	\begin{algorithmic}[1]
		\renewcommand{\algorithmicrequire}{\textbf{Input}:}
		\Require int[] state
		\State Initialize temp = state[0]
		\While{temp $\neq$ 0}
		\For{each s in state}
		\State temp = temp \& s
		\EndFor
		\If{temp $\neq$ 0}
		\State Starting from the highest bit with a value of 1 in temp, assign all the lower bits to 1
		\State belowTemp = temp $>>$ 1, \ aboveTemp = \textasciitilde temp
		\For{each s in state}
		\State below = s \& belowTemp,\ above = s \& aboveTemp
		\State s = below | (above $>>$ 1)
		\EndFor
		\EndIf
		\EndWhile
	\end{algorithmic}
	\label{Alg:delete_line}
\end{algorithm}

This section also involves an operation that can be accelerated: how to set all bits from the highest set bit to the end to 1. For reference, see how a 32-bit integer is set from the highest bit to all subsequent bits to 1 in Java's HashMap. The pseudocode is shown in {\bf{Algorithm \ref{Alg:cal_set_1}}}, with the input being an int-type integer that needs to set all bits from the highest set bit to 1. Since the board's column height is represented by a 32-bit int variable in this paper's design, this algorithm is suitable.

\begin{algorithm}[!htp]
	\floatname{algorithm}{Algorithm}
	\renewcommand{\thealgorithm}{3.3}
	\caption{Set All Bits from the Highest Bit to 1}
	\begin{algorithmic}[1]
		\renewcommand{\algorithmicrequire}{\textbf{Input}:}
		\renewcommand{\algorithmicensure}{\textbf{Output}:}	
		\Require temp
		\State temp = as[i]
		\State temp -= temp $>>$ 1
		\State temp -= temp $>>$ 2
		\State temp -= temp $>>$ 4
		\State temp -= temp $>>$ 8
		\State temp -= temp $>>$ 16
		\Ensure temp
	\end{algorithmic}
	\label{Alg:cal_set_1}
\end{algorithm}

\paragraph{\textbf{GameOver Determination}}
The operation to determine whether the game is over is relatively simple. For a board with a column height of 20, after each piece lands, a variable initialized to 0xf00000 is used to perform a bitwise AND operation on each bit (checking for blocks in rows 21 to 25). If the result of iterating through all columns is not zero, it indicates that blocks have exceeded the 20th row, and the game is over. The pseudocode is shown in {\bf{Algorithm \ref{Alg:is_final}}}, with the input being an int-type array representing the board state and the output being a boolean variable indicating whether the game is over. This algorithm is suitable for a board with a column height of 20; for boards with different column heights, the temp variable should be correspondingly adjusted.

\begin{algorithm}[!htp]
	\floatname{algorithm}{Algorithm}
	\renewcommand{\thealgorithm}{3.4}
	\caption{Determine Game Over}
	\begin{algorithmic}[1]
		\renewcommand{\algorithmicrequire}{\textbf{Input}:}
		\renewcommand{\algorithmicensure}{\textbf{Output}:}	
		\Require Input int[] state
		\State Initialize temp = 0xf00000
		\For{x in each value in state}
		\State temp = temp \& x
		\EndFor
		\If{temp $\neq$ 0}
		\State return True
		\Else
		\State return False
		\EndIf
	\end{algorithmic}
	\label{Alg:is_final}
\end{algorithm}

\subsection{DT Features Based on Bitboard}
Building upon the "afterstate" concept introduced earlier (Fig. \ref{Fig:bitboard_code_case}), an efficient method for computing DT features using a bitboard-based state representation is presented here.To facilitate the calculation of the DT features, this paper does not only save the states of the 10 columns when designing the int array \textit{state} to represent the state of the Tetris game board. The int array \textit{state} is defined as 15-dimensional, where dimensions 0 to 9 represent the states of the 10 columns from left to right on the board; the 10th dimension is \textit{reward}, indicating the number of lines cleared when a piece falls; the 11th dimension is \textit{score}, indicating the cumulative number of lines cleared; the 12th dimension is \textit{randomBlock}, a randomly generated integer ranging from 0 to 6; the 13th dimension is \textit{dropHeight}, representing the height at which a piece falls, and during the piece overlap detection, the height at which the first overlap occurs is \textit{dropHeight}; the 14th dimension is \textit{deleteLine}, indicating the lines to be cleared, calculated as shown in the pseudocode in Algorithm \ref{Alg:deleteLine}. Initially, \textit{state[12]} is initialized to a random integer between 0 and 6, and the rest are initialized to zero. When a piece falls from state $s$ to state $s'$, when calculating the state of the 10 columns in state $s'$, the four values other than the random piece are also calculated and saved in the array, along with the new random piece, denoted as the new array \textit{as}. The \textit{as} array is used in the subsequent calculation of the DT features.

\begin{algorithm}[!htp]
	\floatname{algorithm}{Algorithm}
	\renewcommand{\thealgorithm}{3.5}
	\caption{caculate deleteLine}
	\begin{algorithmic}[1]
		\renewcommand{\algorithmicrequire}{\textbf{Input}:}
		\renewcommand{\algorithmicensure}{\textbf{Output}:}	
		\Require int[] state
		\State deleteLine = state[0]
		\For{i in \{1, 2, $\cdots$, 9\}}
		\State deleteLine = deleteLine \& state[i]
		\EndFor
		\Ensure deleteLine
	\end{algorithmic}
	\label{Alg:deleteLine}
\end{algorithm}

After introducing the state array, we will now explain how the DT features are specifically calculated.
\begin{itemize}
	\item[(i)] \textbf{Landing height}: The height at which the piece falls, $dropHeight$, is already saved in the state array, and this value is denoted as $t_y$. When defining the piece class based on the bitboard design, the height of the piece in this shape has already been stored, so the landing height can be directly calculated. In Figure \ref{Fig:afterstate}, the height at which the piece falls is 1, and the height of the piece in this shape is 3. Therefore, $f_{\text{landing height}} = 1 + (3 - 1)/2 = 2$.
	\item[(ii)] \textbf{Eroded piece cells}: The pseudocode is shown in {\bf{Algorithm \ref{Alg:cal_eroded_piece_cells}}}. The inputs are an int array $as$ and the current falling piece $b$, where the Piece class is the piece class proposed before. The algorithm uses short-circuit programming: if $as[10]$ (i.e., reward) is 0, it is clear that no lines have been cleared, so it directly returns 0. $deleteLine$ stores the cleared lines. Since the piece is stored aligned with the first row, $deleteLine$ is first shifted downward by $dropHeight$ to align with the piece. Then, a bitwise AND operation is performed on each column of the piece, and the number of 1s is counted. In Figure \ref{Fig:afterstate}, the number of cleared lines is 2, and the number of cells in the piece that participated in the clearing is 3. Therefore, $f_{\text{eroded piece cells}} = 2 \times 3 = 6$.
	\begin{algorithm}[!htb]
		\floatname{algorithm}{Algorithm}
		\renewcommand{\thealgorithm}{3.6}
		\caption{Calculating the Eroded piece cells feature}
		\begin{algorithmic}[1]
			\renewcommand{\algorithmicrequire}{\textbf{Input}:}
			\renewcommand{\algorithmicensure}{\textbf{Output}:}   
			\Require int[] $as$, Piece $p$
			\If{$as[10] == 0$}
			\State return 0
			\EndIf
			\State $deleteLine = as[14] >>> as[13]$, $num = 0$
			\For{$i$ in \{0, $\cdots$, piece.width\}}
			\State $temp = deleteLine \& p.piece[i]$
			\While{$temp \neq 0$}
			\State $temp \&= (temp - 1)$
			\State $++num$
			\EndWhile
			\EndFor
			\Ensure $num \times as[10]$
		\end{algorithmic}
		\label{Alg:cal_eroded_piece_cells}
	\end{algorithm}
	
	\item[(iii)] \textbf{Row transitions}: The state array stores the states of 10 columns. According to the formal definition, it is divided into three parts for calculation: counting the transitions between each column from the first to the ninth column and the next column, the number of empty cells in the first column, and the number of empty cells in the tenth column. The pseudocode is shown in {\bf{Algorithm \ref{Alg:cal_row_transitions}}}. For a column-based Tetris game, 0x3ff should be replaced with 0xfffff. In Figure \ref{Fig:afterstate}, the number of empty cells in the first column is 5, the number of empty cells in the tenth column is 6, and the number of cell transitions between the first to ninth columns and the next column are \{0, 4, 2, 0, 2, 1, 2, 0, 0\}, respectively. Therefore, $f_{\text{row transitions}} = 5 + 6 + (4 + 2 + 2 + 1 + 2) = 22$.
	\begin{algorithm}[!htb]
		\floatname{algorithm}{Algorithm}
		\renewcommand{\thealgorithm}{3.7}
		\caption{Calculating the Row transitions feature}
		\begin{algorithmic}[1]
			\renewcommand{\algorithmicrequire}{\textbf{Input}:}
			\renewcommand{\algorithmicensure}{\textbf{Output}:}   
			\Require int[] $as$
			\State $num = 0$, $y = as[0] \oplus \ 0x3ff$
			\While {$y \neq 0$}
			\State $y = y \& (y-1)$
			\State $++num$
			\EndWhile
			\State $y = as[9] \oplus \ 0x3ff$
			\While {$y \neq 0$}
			\State $y = y \& (y-1)$
			\State $++num$
			\EndWhile
			\For{$i$ in \{0, $\cdots$, 8\}}
			\State $y = as[i] \oplus \ as[i+1]$
			\While{$y \neq 0$}
			\State $y \&= (y - 1)$
			\State $++num$
			\EndWhile
			\EndFor
			\Ensure $num$
		\end{algorithmic}
		\label{Alg:cal_row_transitions}
	\end{algorithm}
	
	\item[(iv)] \textbf{Column transitions}: The pseudocode is shown in {\bf{Algorithm \ref{Alg:cal_column_transitions}}}. By shifting the column up one cell using $as[i] << 1$, and adding 1 because the bottom of the board is considered to have a block, the state of the first row is set to have a block after the shift. The XOR operation can then be used to count the transitions in each column. In Figure \ref{Fig:afterstate}, the number of transitions in each column from the first to the tenth column is \{1, 1, 3, 1, 1, 1, 1, 1, 1, 1\}, respectively. Therefore, $f_{\text{column transitions}} = 1 + 1 + 3 + 1 + 1 + 1 + 1 + 1 + 1 + 1 = 12$.
	\begin{algorithm}[!htb]
		\floatname{algorithm}{Algorithm}
		\renewcommand{\thealgorithm}{3.8}
		\caption{Calculating the Column transitions feature}
		\begin{algorithmic}[1]
			\renewcommand{\algorithmicrequire}{\textbf{Input}:}
			\renewcommand{\algorithmicensure}{\textbf{Output}:}   
			\Require int[] $as$
			\State $num = 0$
			\For{$i$ in \{0, $\cdots$, 9\}}
			\State $y = as[i] \oplus \ ((as[i] << 1) + 1)$
			\While {$y \neq 0$}
			\State $y = y \& (y-1)$
			\State $++num$
			\EndWhile
			\EndFor
			\Ensure $num$
		\end{algorithmic}
		\label{Alg:cal_column_transitions}
	\end{algorithm}
	\item[(v)] \textbf{Holes}: The pseudocode is shown in {\bf{Algorithm \ref{Alg:cal_holes}}}. For each column, starting from the highest 1 and filling all positions below it with 1, the number of holes is counted by XORing with the original column state and counting the number of 1s. Clearly, in Figure \ref{Fig:afterstate}, the number of holes is 1.
	
	\begin{algorithm}[!htb]
		\floatname{algorithm}{Algorithm}
		\renewcommand{\thealgorithm}{3.9}
		\caption{Calculating the Holes feature}
		\begin{algorithmic}[1]
			\renewcommand{\algorithmicrequire}{\textbf{Input}:}
			\renewcommand{\algorithmicensure}{\textbf{Output}:}   
			\Require int[] $as$
			\State $num = 0$
			\For{$i$ in \{0, $\cdots$, 9\}}
			\If{$as[i] > 1$}
			\State $y = as[i]$
			\State $y |= y >> 1$
			\State $y |= y >> 2$
			\State $y |= y >> 4$
			\State $y |= y >> 8$
			\State $y |= y >> 16$
			\State $y = y \oplus \ as[i]$
			\While {$y \neq 0$}
			\State $y = y \& (y-1)$
			\State $++num$
			\EndWhile
			\EndIf
			\EndFor
			\Ensure $num$
		\end{algorithmic}
		\label{Alg:cal_holes}
	\end{algorithm}
	
	\item[(vi)] \textbf{Board well}: The pseudocode is shown in {\bf{Algorithm \ref{Alg:cal_board_wells}}}. First, a new array $exState$ is constructed, where the first and twelfth dimensions are all 1 (if the column height is 20, replace 0x3ff with 0xfffff), and the second to eleventh dimensions are the int values representing the ten-column states. Then, each column is traversed to calculate the cumulative depth of each well. In Figure \ref{Fig:afterstate}, the cumulative well depths for each column are \{0, 0, 1 + 1, 0, 0, 1, 0, 0, 0, 0\}. Therefore, $f_{\text{board wells}} = 2 + 1 = 3$.
	\begin{algorithm}[!htb]
		\floatname{algorithm}{Algorithm}
		\renewcommand{\thealgorithm}{3.10}
		\caption{Calculating the Board well feature}
		\begin{algorithmic}[1]
			\renewcommand{\algorithmicrequire}{\textbf{Input}:}
			\renewcommand{\algorithmicensure}{\textbf{Output}:}   
			\Require int[] $as$
			\State $num = 0$, int[] $exState = new int[12]$
			\State $exState[0] = exState[11] = 0x3ff$
			\For{$i$ in \{1, $\cdots$, 10\}}
			\State $exState[i] = state[i-1]$
			\EndFor
			\For{$i$ in \{1, $\cdots$, 10\}}
			\For{$j$ in \{0, $\cdots$, 9\}}
			\If{((exState[i] $>>$ j) \& 1) == 0}
			\State $++y$
			\If{((exState[i - 1] $>>$ j) \& 1) == 1 and ((exState[i + 1] $>>$ j) \& 1) == 1}
			\State $num = num + y$
			\EndIf
			\Else
			\State $y = 0$
			\EndIf
			\EndFor
			\State $y = 0$
			\EndFor
			\Ensure $num$
		\end{algorithmic}
		\label{Alg:cal_board_wells}
	\end{algorithm}
	\item[(vii)] \textbf{Hole depth}: The pseudocode is shown in {\bf{Algorithm \ref{Alg:cal_hole_depth}}}. For each column, starting from the bottom and moving upwards, the first hole is found, and then the number of cells above the hole that contain blocks is counted. In Figure \ref{Fig:afterstate}, there is only one hole in the third column with one block above it. Therefore, $f_{\text{hole depth}} = 1$.
	\begin{algorithm}[!htb]
		\floatname{algorithm}{Algorithm}
		\renewcommand{\thealgorithm}{3.11}
		\caption{Calculating the Hole depth feature}
		\begin{algorithmic}[1]
			\renewcommand{\algorithmicrequire}{\textbf{Input}:}
			\renewcommand{\algorithmicensure}{\textbf{Output}:}   
			\Require int[] $as$
			\State $num = 0$
			\For{$i$ in \{0, $\cdots$, 9\}}
			\State $y = \oplus (as[i] \oplus \ (as[i] + 1))$
			\State $y = as[i] \& y$
			\While {$y \neq 0$}
			\State $y = y \& (y-1)$
			\State $++num$
			\EndWhile
			\EndFor
			\Ensure $num$
		\end{algorithmic}
		\label{Alg:cal_hole_depth}
	\end{algorithm}
	\item[(viii)] \textbf{Rows with holes}: The pseudocode is shown in {\bf{Algorithm \ref{Alg:cal_rowHoles}}}. Each column is traversed to find rows with empty cells, which are stored in $y2$. Finally, the total number of rows with empty cells in $y2$ is counted. In Figure \ref{Fig:afterstate}, there is only one row with a hole. Therefore, $f_{\text{rows with holes}} = 1$.
	\begin{algorithm}[!htb]
		\floatname{algorithm}{Algorithm}
		\renewcommand{\thealgorithm}{3.12}
		\caption{Calculating the Row with holes feature}
		\begin{algorithmic}[1]
			\renewcommand{\algorithmicrequire}{\textbf{Input}:}
			\renewcommand{\algorithmicensure}{\textbf{Output}:}   
			\Require int[] $as$
			\State $num = 0$, $y2 = 0$
			\For{$i$ in \{0, $\cdots$, 9\}}
			\If{$as[i] > 1$}
			\State $y1 = as[i]$
			\State $y1 |= y1 >> 1$
			\State $y1 |= y1 >> 2$
			\State $y1 |= y1 >> 4$
			\State $y1 |= y1 >> 8$
			\State $y1 |= y1 >> 16$
			\State $y1 = y1 \oplus \ as[i]$
			\State $y2 = y2 |= y1$
			\EndIf
			\EndFor
			\While {$y2 \neq 0$}
			\State $y2 = y2 \& (y2-1)$
			\State $++num$
			\EndWhile
			\Ensure $num$
		\end{algorithmic}
		\label{Alg:cal_rowHoles}
	\end{algorithm}
	\item[(ix)] \textbf{Pattern diversity}: When calculating the height differences between adjacent columns, the height of each column needs to be calculated. This can be done using {\bf{Algorithm \ref{Alg:cal_height}}}, which is denoted as the $searchHeight$ function. The pseudocode for calculating Pattern diversity is shown in {\bf{Algorithm \ref{Alg:cal_pattern}}}. Since $searchHeight$ returns the column height minus 1, for a column with no blocks, to align with the result of $searchHeight$, the column height is set to -1. In Figure \ref{Fig:afterstate}, the height differences between each column and the next column are \{0, 3, -1, 0, 2, -1, -2, 0, 0\}. The unique values are \{-2, -1, 0, 2\}. Therefore, $f_{\text{pattern diversity}} = 4$.
	\begin{algorithm}[!htb]
		\floatname{algorithm}{Algorithm}
		\renewcommand{\thealgorithm}{3.13}
		\caption{Calculating the Pattern diversity feature}
		\begin{algorithmic}[1]
			\renewcommand{\algorithmicrequire}{\textbf{Input}:}
			\renewcommand{\algorithmicensure}{\textbf{Output}:}   
			\Require int[] $as$
			\State int[] $h = new int[10]$, int[] $count = \{0, 0, 0, 0, 0\}$, int[] $p = \{-2, -1, 0, 1, 2\}$, int $num = 0$
			\For{$i$ in \{0, $\cdots$, 9\}}
			\If{$as[i] \neq 0$}
			\State $h[i] = searchHeight(as[i])$
			\Else
			\State $h[i] = -1$
			\EndIf
			\EndFor
			\For{$i$ in \{0, $\cdots$, 8\}}
			\State $h1 = h[i] - h[i+1]$
			\For{$j$ in \{0, $\cdots$, 4\}}
			\If{$h1 == p[j]$}
			\State $++count[j]$
			\EndIf
			\EndFor
			\EndFor
			\For{$i$ in \{0, $\cdots$, 4\}}
			\If{$count[i] \neq 0$}
			\State $++num$
			\EndIf
			\EndFor
			\Ensure $num$
		\end{algorithmic}
		\label{Alg:cal_pattern}
	\end{algorithm}
\end{itemize}

\subsection{Python interface}
In the field of AI today, deep reinforcement learning (DRL) algorithms based on neural networks have demonstrated significant potential in game development and training. Although the game environment is implemented in Java, Python offers unparalleled advantages for AI training. The rise of open-source frameworks (e.g., PyTorch and TensorFlow) has equipped AI developers with powerful and user-friendly API interfaces. For example, PyTorch allows for easy implementation of the required network structures, and gradient calculations via backpropagation can be performed with just a few API calls. Therefore, the final solution adopted in this paper is as follows: The bitboard-based Tetris game environment is implemented in Java, while each operation is invoked through Python, utilizing the open-source library Jpype \cite{nelson2020jpype} version 1.5.0. Jpype is a tool for invoking Java code within the Python environment, enabling Python programs to directly call Java classes and their methods. JPype supports data type conversions between Python and Java (e.g., converting a Python list to a Java array) and enables the direct import of trained weights into the Java-implemented bitboard-based Tetris game for testing. This approach leverages the performance advantages of the Java language while taking full advantage of the convenience of Python in AI development.
For ease of use, a Tetris class is implemented in Python following the OpenAI Gym standard interface. This class is globally unique and initializes the Java Virtual Machine (JVM) and the Java-implemented classes upon initialization. The JDK version used by the JVM is consistent with the version used to compile the Java classes, both using openjdk-21.0.1. The following API interfaces are exposed and provided in Python:

\begin{itemize}
	\item[(a)] \texttt{Tetris.reset()}: Initializes the Tetris game environment and returns a 15-dimensional integer array \texttt{state}. The definition of the \texttt{state} array is the same as described in Section before.
	
	\item[(b)] \texttt{Tetris.isFinal()}: Takes \texttt{state} as input and returns a boolean variable indicating whether the game is over.
	
	\item[(c)] \texttt{Tetris.parallel\_episode()}: Takes the linear weights of the network as input, tests 10,000 games in parallel in Java, and returns the average score of these games. This method is only applicable to the 9-dimensional weights trained with DT features.
	
	\item[(d)] \texttt{Tetris.get\_9feature()}: Takes \texttt{state} as input and returns the DT features and mask for up to 34 possible successor states based on the given \texttt{state}.
	
	\item[(e)] \texttt{Tetris.step()}: Takes \texttt{state} and an action \texttt{a} as input, and returns the next state array, the reward for the action, and a boolean indicating whether the game is over.
\end{itemize}

\section{Buffer-based PPO for Tetris AI}

The previous section implemented the Tetris game using bitboard, aiming to achieve significant acceleration over traditional grid-based storage and resolve its computational bottlenecks. To train high-performance agents on the bitboard-based implementation, we designed a buffer-based PPO algorithm that evaluates afterstates, which can rapidly train a well-performing agent. Given the excessive training time required for standard $10\times20$ boards and the prevalent use of more challenging $10\times10$ mini-boards in current research, this section adopts $10\times10$ boards for experiments to balance training efficiency and research representativeness.

\subsection{Afterstate-Based Actor Network}

In the field of reinforcement learning, the Actor network is a key component in policy gradient-based algorithms, whose main function is to learn and output the probability distribution of taking each action under a given state.In Tetris, environmental randomness primarily stems from the generation of the next block type, whereas the placement of the current block and its impact on the board topology exhibit deterministic transition characteristics after an action is executed. Compared to directly estimating the action-value function \(Q(s, a)\), value evaluation based on afterstates can explicitly decouple the deterministic outcome of the agent’s decision from the stochastic disturbances of the environment system. From the theoretical perspective of policy gradients, the traditional \(Q(s, a)\) implicitly requires taking expectations over all subsequent states corresponding to random blocks during updates, which can introduce substantial estimation variance when sampling is insufficient. Since the mapping \(s \times a \to as\) is deterministic, the afterstate value function \(V(as)\) can directly leverage deterministic board features for updates, with expectation calculations applied only to the long-term stochastic distribution of subsequent events. This decoupling mechanism effectively reduces the search space of the value function, significantly lowers the variance of gradient estimates, and thereby enhances the stability of policy updates. This property is particularly crucial in the Tetris task. The line clears and sudden changes in the board state resulting from block placement exhibit highly regular geometric patterns. The Actor network based on afterstates shifts its learning focus from 'evaluating the quality of action-block combinations' to 'evaluating the quality of board configurations.' This not only simplifies the difficulty of feature mapping but also enables the network to achieve deep modeling of board structures with a smaller parameter scale, significantly improving sample utilization efficiency, which will be comparatively analyzed and validated in the experiments presented in the next section.

To effectively harness the parallel computing capabilities of GPUs for the Actor implementation in Tetris, specific optimizations tailored to the game's characteristics are essential. For instance, the Actor network evaluating afterstates shares similarities with those assessing action-value functions. Typically, an Actor network takes a state $s$ or its features as input and outputs the probability distribution of actions $a$ under $s$. However, in Tetris, the input is uniquely processed: it comprises afterstate features for all feasible actions $a$ from state $s$, preprocessed with feature extraction, concatenation, and a mask to enhance action selection based on afterstate data. The key steps are:

\begin{enumerate}
	\item[(i)] \textbf{Feature extraction}: Compute afterstate features for each feasible action (e.g., 9 for the O-piece), with infeasible actions assigned arbitrary values (eliminated by the mask, thus irrelevant to softmax output). This chapter uses DT features to quantify game progress.
	
	\item[(ii)] \textbf{Feature concatenation}: Combine afterstate features into a 306-dimensional input vector $f(s,a)_{\text{all}}$, accommodating up to 34 actions, with masking for infeasible actions varying by piece (9 to 34).
	
	\item[(iii)] \textbf{Masking mechanism}: Apply a mask (1 for feasible, 0 for infeasible) to set infeasible action probabilities to 0 via negative infinity, ensuring only feasible actions are considered.
\end{enumerate}

\begin{figure}[!htp]
	\centering
	\includegraphics[width=0.7\linewidth]{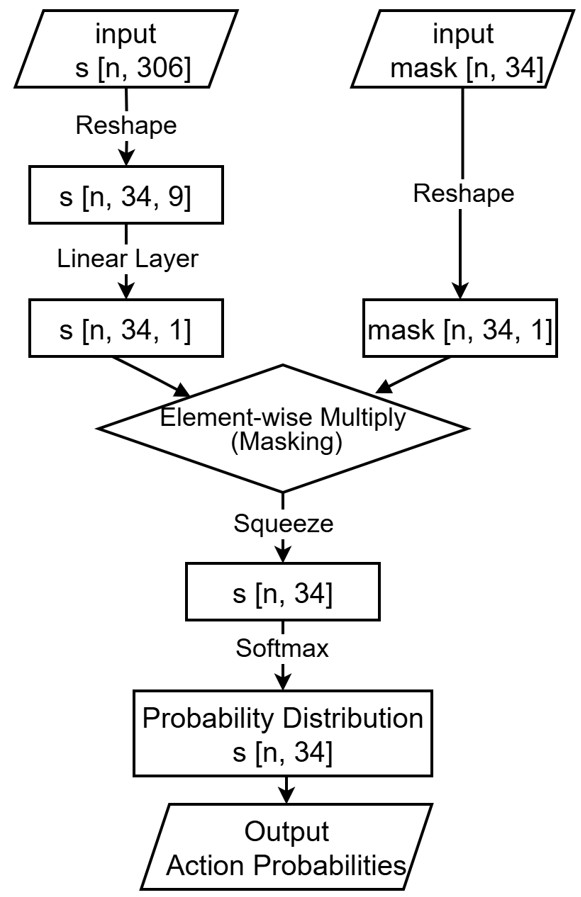}
	\caption{The feedforward process of the Actor evaluating afterstates}
	\label{Fig:Actor}
\end{figure}

The Actor structure for afterstate evaluation is depicted in Fig.\ref{Fig:Actor}, where input is a 306-dimensional feature and $n (\geq 1)$ is the batch size. For in-game decisions, only current afterstates are input ($n=1$); for policy updates, batch processing uses trajectory or buffer data. To boost sampling efficiency and utilize GPU parallelism, CPU handles game sampling, while GPU manages agent updates. Through reshape operations, a shared linear layer evaluates all afterstates, masks out infeasible actions, and applies softmax to yield action probabilities.

\subsection{Buffer-Based PPO}

In reinforcement learning, particularly for long-episode games like Tetris, training efficiency critically depends on balancing sampling and policy update times. Standard trajectory-based PPO---which collects a full episode before performing updates---suffers from a severe imbalance: sampling dominates computation, while policy updates consume negligible effort.
To address this, we propose a Buffer-based PPO algorithm to mitigate the time disparity. The core idea is to adjust the training pace, shifting from the trajectory-based PPO's approach of waiting for a full game before updating, to initiating training once the Buffer collects a batchSize number of samples. During each training cycle, the process loops epoch times, randomly sampling miniBatchSize data from the Buffer for training. This method aligns the frequency of sampling and policy updates, enhancing the agent's training speed. The pseudocode for the Buffer-based PPO algorithm is provided in {\bf{Algorithm \ref{Alg:PPO_buffer}}}.

\begin{algorithm}[!htp]
	\floatname{algorithm}{Algorithm}
	\renewcommand{\thealgorithm}{4.1}
	\caption{Buffer-based PPO Algorithm}
	\footnotesize
	\begin{algorithmic}[1]
		\State Initialize Actor, Critic networks, and Replay Buffer, set step = 0
		\While{step $\leq$ totalStep}
		\State done = false, $s$ = env.reset(), set Buffer.size = 0
		\While{not done}
		\State Compute $f(s,a)_{\text{all}}$ and $mask$ based on state $s$
		\State Actor selects action $a$ using $f(s,a)_{\text{all}}$ and $mask$
		\State Execute env.step($a$), receive next state $s'$ and reward $r$
		\State Store transition data in Buffer for training
		\If{Buffer.size == $batchSize$}
		\For{each training epoch $k$}
		\State Randomly sample a $miniBatchSize$ from Buffer
		\State Compute TD error: 
		\State \quad$\delta_t = r + \gamma \cdot \text{Critic}(f(s_t, a_t)) - \text{Critic}(f(s_t))$
		\State Compute Generalized Advantage Estimation:
		\State \quad $A_t = \sum_{l=0}^{\infty} (\gamma \lambda)^l \delta_{t+l}$
		\State Update Actor:
		\State \quad $\theta \gets \theta + \alpha_\theta \nabla_\theta L_{\text{a}}$
		\State \quad where $L_{\text{a}} = \min\left[ \frac{\pi_{\theta}(a|s)}{\pi_{\theta_k}(a|s)} A_t, \right.$
		\State \quad \quad $\left. \text{clip}\left( \frac{\pi_{\theta}(a|s)}{\pi_{\theta_k}(a|s)}, 1-\epsilon, 1+\epsilon \right) A_t \right]$
		\State Update Critic:
		\State \quad $w \gets w + \alpha_w \sum_{t} \delta_t \nabla_w V_w(s_t)$
		\EndFor
		\State Reset Buffer.size = 0
		\EndIf
		\EndWhile
		\EndWhile
	\end{algorithmic}
	\normalsize
	\label{Alg:PPO_buffer}
\end{algorithm}

With the update shifting from episode-based to Buffer-based, the learning rate decay is adjusted to linear decay, as shown in Eq.\eqref{eq:lr_decay}, applied to both Actor and Critic networks. This approach suits Buffer-based training, offering a high initial learning rate to speed up early parameter updates, then gradually reducing it to stabilize updates, prevent oscillations, and enhance convergence to an optimal solution.

\begin{equation}\label{eq:lr_decay}
	\alpha = \alpha_{\text{init}} \times \left(1 - \frac{\text{currentStep}}{\text{totalSteps}}\right)
\end{equation}
\section{Experiments}

In this section, we present a series of experiments designed to validate the proposed bitboard-based Tetris implementation and evaluate the performance of various policy optimization approaches. First, we assess the correctness and efficiency of the bitboard implementation by replicating prior benchmarks and comparing execution speeds. Subsequently, we clarified the criteria for selecting evaluation targets and learning algorithms. At the evaluation level, both action-value and afterstate representations were initially identified as core indicators, while at the algorithmic level, basic reinforcement learning and Proximal Policy Optimization (PPO) were compared in parallel. Guided by a greedy selection strategy that prioritizes higher-performing configurations, the experimental results demonstrated that afterstate evaluation provided superior effectiveness over action-value estimation, leading to its adoption as the unified evaluation indicator in subsequent experiments. Similarly, PPO outperformed basic reinforcement learning and was established as the core algorithmic framework.
Building upon these selections, further optimization and refinement were pursued. Beginning with conventional trajectory-based PPO, we introduced a buffer mechanism to enhance sample utilization and update efficiency, ultimately forming an improved buffered PPO algorithm that achieved a better balance between learning stability and computational efficiency.

\subsection{Validation of Bitboard-Based Implementation on Tetris}

\paragraph{\textbf{Benchmark Replication for Correctness}}
Victor Gabillon et al. \cite{gabillon2013approximate} achieved an average score of 4200 using the CBMPI algorithm and DT features on a 10$\times$10 board, with the best training run scoring 5200. Gabillon et al. provided two sets of weights, DT-10 and DT-20, in their paper, as shown in Table \ref{table_dt1020}. Gabillon tested each set of weights over 10,000 games. The DT-10 weights achieved an average score of approximately 5000 on the 10$\times$10 board and about 29,000,000 on the 10$\times$20 board. The DT-20 weights achieved an average score of approximately 4300 on the 10$\times$10 board and about 51,000,000 on the 10$\times$20 board.

This paper also tests based on these two sets of weights to verify the correctness of the game environment. On the 10$\times$10 board, 10,000 games were tested for each set of weights, and the results are shown in Table \ref{table:test_weight}. DT-10 and DT-20 directly select the action with the highest linear sum of DT feature values based on all subsequent states' features $f(s,a)$, consistent with the method in the original paper. The test results are consistent with the conclusions given in the original paper.

On the 10$\times$20 board, 10 games were tested for each set of weights, and the results are shown in Table \ref{table:test_weight}, verifying that the Tetris game design based on bitboard in this paper is feasible on both 10$\times$10 and 10$\times$20 boards.
\begin{table}[htb]
	\centering
	\caption{Scores with different weights on different board sizes
		(Mean(SD))}
	\label{table:test_weight}
	\resizebox{0.45\textwidth}{!}{%
		\begin{tabular}{@{}l l c c@{}}
			\toprule
			\textbf{Board Size} & \textbf{Weights} & \textbf{Gabillon et al.~(2013)\cite{gabillon2013approximate}} & \textbf{Our bitboard impl} \\
			\midrule
			\multirow{2}{*}{10$\times$10(10,000 games)} & DT-10 & 5,000(-) & 5,152.02(5,137.51) \\
			& DT-20 & 4,300(-) & 4,188.61(4,153.42) \\
			\midrule
			\multirow{2}{*}{10$\times$20(10 games)} & DT-10 & 29,000,000(-) & 54,377,543(54,353,156) \\
			& DT-20 & 51,000,000(-) & 75,093,204(65,000,224) \\
			\bottomrule
	\end{tabular}}
\end{table}

\paragraph{\textbf{Runtime Efficiency Comparison}}
A comparison is made between the open-source \texttt{gym-tetris} implementation and our bitboard-based Tetris implementation, described in detail below:

\begin{itemize}
	\item [(i)] \textbf{Standard open-source implementation}: \texttt{gym-tetris} \cite{gym-tetris} provides an OpenAI Gym interface based on the \texttt{nes-py} emulator for playing Tetris on the Nintendo Entertainment System (NES). This implementation supports reinforcement learning research and multiple game modes with different reward mechanisms. Two main modes are available: A-type (standard endurance Tetris) and B-type (arcade-style, requiring a set number of lines to win). Each mode includes four variants (v0–v3) with different reward structures (e.g., scoring, line-clear bonuses, or penalties for board height). In our experiments, we use the endurance-type Tetris with a line-clear reward function, corresponding to the \texttt{TetrisA-v1} version. Although the library supports rendering, image rendering is disabled during training due to high computational cost.
	\item [(ii)] \textbf{Tetris based on bitboard}: To maximize performance, we implemented Tetris using bitboards. Considering the efficiency limitations of Python for bitwise operations, this implementation was completed in Java, ensuring high execution speed and stability.
\end{itemize}

For performance testing, each implementation was sampled for 10,000 steps, with the agent selecting actions randomly to avoid time overhead from decision-making. The experiments were conducted on a Windows 11 machine with an AMD R7-7735H CPU and 16GB of RAM, which was used for all subsequent tests. The time comparison of the three implementations is shown in Table \ref{table:env_time}. It is evident that the bitboard-based Tetris implementation has a significant speed advantage. For 10,000 samples, it runs approximately 53 times faster than the standard OpenAI Gym implementation. Even when called via Python, it achieves around 22 times the speed of the standard implementation, despite some performance loss due to thread communication.
\begin{table}[htb]
	\centering
	\caption{Time comparison of Tetris implementations for 10,000 samples}
	\label{table:env_time}
	\resizebox{0.45\textwidth}{!}{%
		\begin{tabular}{@{}lc@{}}
			\toprule
			\textbf{Implementation} & \textbf{Time for 10,000 samples} \\ \midrule
			OpenAI Gym\cite{gym-tetris} & 12.92 seconds \\
			Bitboard Tetris (ours) & \textbf{0.24 seconds} \\
			Bitboard Tetris via Python (ours) & \textbf{0.59 seconds} \\
			\bottomrule
	\end{tabular}}
\end{table}

\subsection{Evaluation of PPO Variants for Tetris Agents}
\paragraph{\textbf{Assessment of Afterstate vs. Action-Value}}
To validate the afterstate evaluating Actor, experimental tests are conducted in this section. The REINFORCE algorithm is employed to eliminate the influence of the Critic network. Additionally, to prevent performance discrepancies arising from differences in neural network architectures, only linear layers with dimensionality identical to the input vector are utilized.To accelerate the convergence of the $softmax$ function, an annealing factor $\tau$ is used in the experiments to speed up convergence. The values input to the $softmax$ function are first divided by $\tau$. Here, $\tau(i)$ represents the temperature at the $i$-th episode, controlling the balance between exploitation and exploration. $\tau_0$ denotes the initial temperature, and $\tau_k$ controls the annealing rate.

\begin{equation}
	\tau(i) = \frac{\tau_0}{1 + \tau_k \times i}
\end{equation}

\begin{figure}[!htp]
	\centering
	\includegraphics[width=1\linewidth]{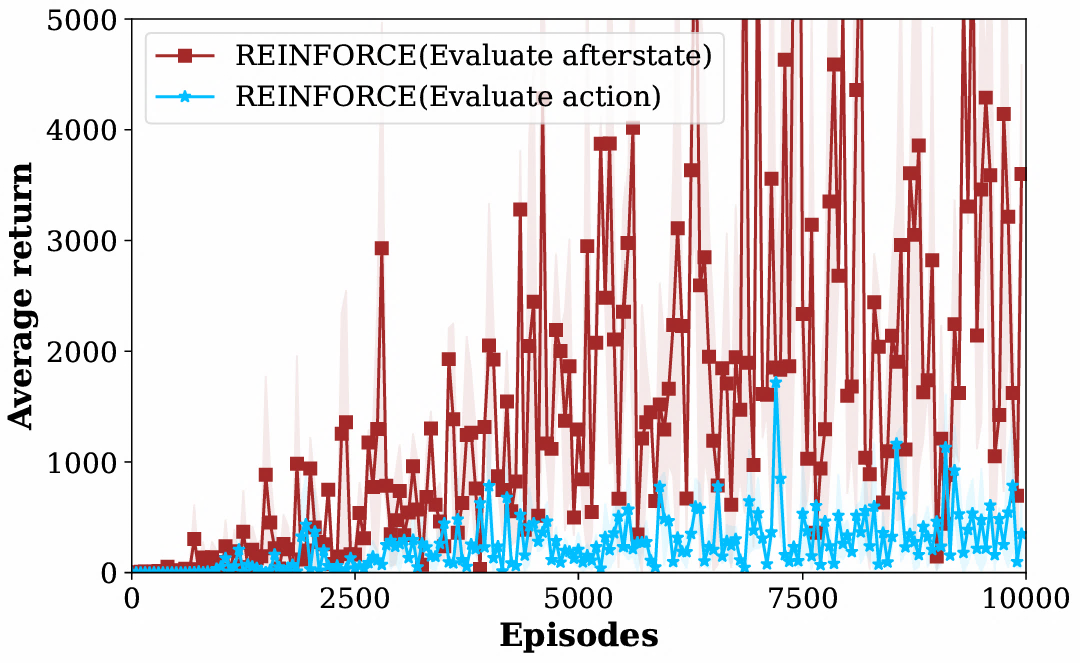}
	\caption{Comparison of Action-Value Evaluating Actor and Afterstate Evaluating Actor}
	\label{Fig:reinforce}
\end{figure}

The parameters include a discount factor of $\gamma = 1$ (undiscounted returns), an initial temperature of $\tau_0 = 0.5$, an annealing rate of $\tau_k = 0.00025$, and a learning rate of $\alpha = 1e-3$. For the comparison, we evaluate two models: the action-value Actor with a 48-dimensional input and the afterstate Actor with a 9-dimensional input, both utilizing linear layers to avoid architecture bias. Due to the long training time of the REINFORCE algorithm, the experiment is conducted over two independent trials. The final results are presented as a mean-variance plot in Figure \ref{Fig:reinforce}.

The experimental results indicate that the Actor evaluating afterstates utilizes fewer weights and achieves significantly better performance than the Actor evaluating the action-value function. Analyzing the reasons for this, the placement of a piece can trigger drastic changes in the spatial structure, creating state transitions far more complex than those assumed by traditional models. Although the DT features numerically represent the current board state, they must contend with dynamic changes (e.g., line-clear cascades and spatial reconfiguration) that occur upon tetromino placement. The traditional method of evaluating action-value, due to its lack of a targeted environmental modeling mechanism, struggles to capture the patterns of state evolution in real-time, leading to delayed policy updates. Although theoretically the expressive power could be enhanced by stacking more network layers, this "brute-force" solution not only significantly increases computational costs but is also highly prone to overfitting, and thus is not a reasonable path for algorithm optimization. Therefore, in subsequent experiments of this paper, the Actor that evaluates afterstates is selected.

\paragraph{\textbf{Performance of Trajectory-Based PPO Baseline}}
The Trajectory-Based PPO implementation serves as the primary baseline, representing the standard on-policy training paradigm commonly used in reinforcement learning. The experimental parameter settings are shown in Table \ref{Tab:ppo_trajectory_hp}.

\begin{table}[!htb]
	\centering
	\caption{Trajectory-Based PPO hyperparameters}
	\label{Tab:ppo_trajectory_hp}
	\resizebox{0.45\textwidth}{!}{
		\begin{tabular}{@{} l c l c l c @{}}
			\toprule
			\textbf{Param} & \textbf{Value} & \textbf{Param} & \textbf{Value} & \textbf{Param} & \textbf{Value} \\
			\midrule
			$\gamma$           & 0.99           & $\tau_0$           & 0.5            & $\alpha_\theta$    & 3e-4 \\
			$\lambda$          & 0.99           & $\tau_k$           & 0.00025        & $\alpha_\omega$    & 3e-4 \\
			$epoch$            & 10             & $\epsilon$         & 0.1            & $episode$          & 12,500 \\
			\bottomrule
	\end{tabular}}
\end{table}

The experiment was independently repeated 5 times to reduce randomness, and the final results are plotted as the mean-variance graph in Fig.\ref{Fig:PPO_trajectory}:

\begin{figure}[!htp]
	\centering
	\includegraphics[width=1\linewidth]{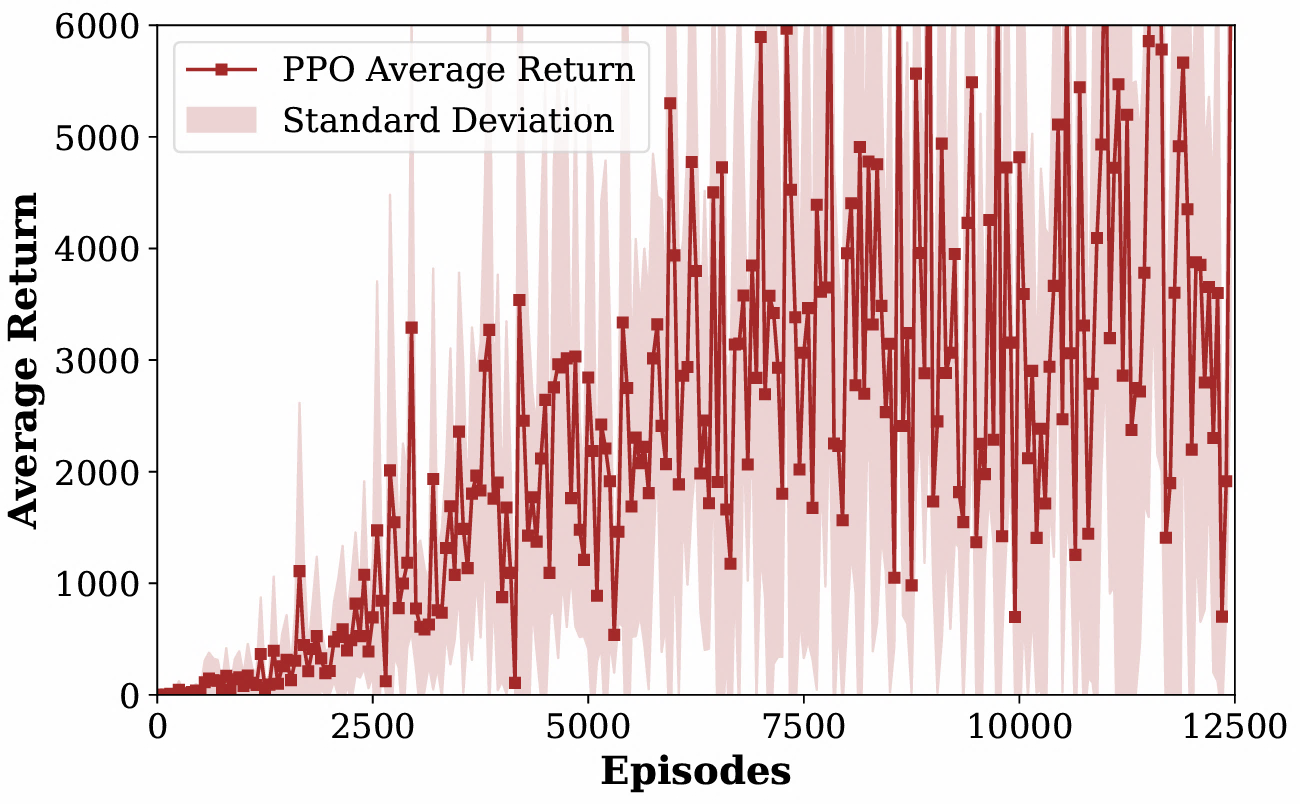}
	\caption{Trajectory-Based PPO Algorithm}
	\label{Fig:PPO_trajectory}
\end{figure}

As can be seen from the figure, the trajectory-based PPO algorithm achieves significant results when training Tetris game agents. As the number of training rounds increases, the average reward gradually rises. The average value of the rewards from the last 150 episodes of five independent training runs is 3840.30, which is taken as the average convergence value of the training. By combining the PPO algorithm with the Actor that evaluates afterstates, results close to those of CBMPI are achieved.

\paragraph{\textbf{Efficiency Gains with Buffer-Based PPO}}
The experimental parameter settings are shown in Table \ref{Tab:ppo_buffer_hp}.

\begin{table}[htb]
	\centering
	\caption{Buffer-based PPO hyperparameters}
	\label{Tab:ppo_buffer_hp}
	\resizebox{0.45\textwidth}{!}{
		\begin{tabular}{@{} l c  l c  l c @{}}
			\toprule
			\textbf{Parameter} & \textbf{Value} & \textbf{Parameter} & \textbf{Value} & \textbf{Parameter} & \textbf{Value} \\
			\midrule
			$\gamma$           & 0.99           & $\epsilon$         & 0.2            & $batchSize$        & 2,048 \\
			$\lambda$          & 0.99           & $\alpha_\theta$    & 3e-4           & $miniBatchSize$    & 256  \\
			$epoch$            & 10             & $\alpha_\omega$    & 3e-4           & $totalSteps$       & 61,440 \\
			\bottomrule
	\end{tabular}}
\end{table}

\begin{figure}[!htp]
	\centering
	\includegraphics[width=1\linewidth]{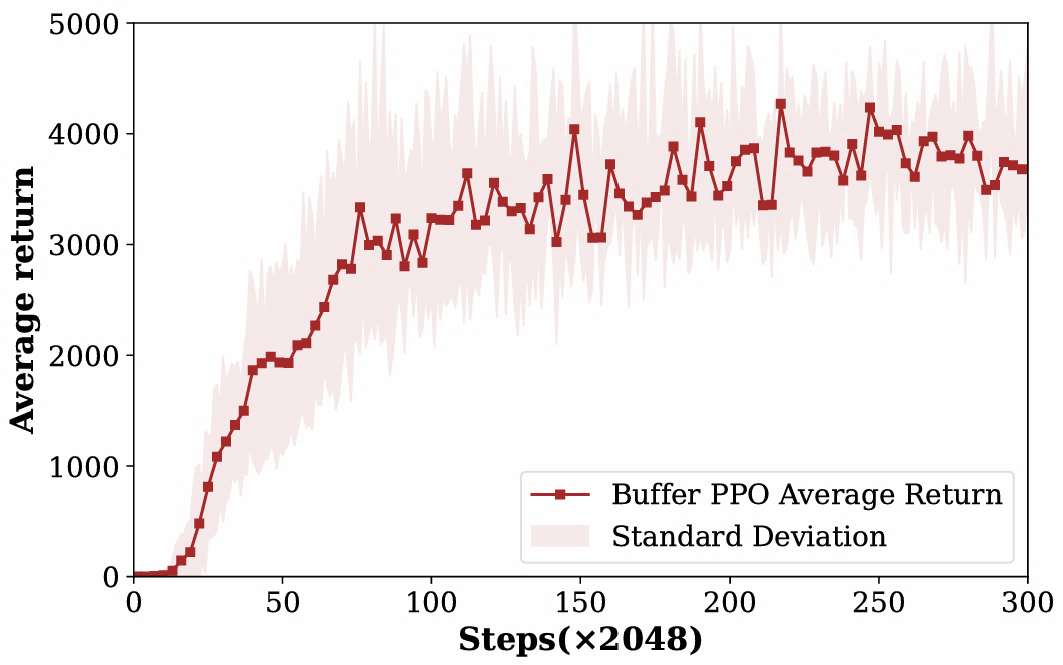}
	\caption{Buffer-Based PPO Algorithm}
	\label{Fig:PPO_buffer}
\end{figure}

In the previous trajectory-based PPO algorithm, the mean-variance plot is drawn using the return per episode as the evaluation metric during training. For the Buffer-based PPO algorithm, the agent is tested for 50 episodes after each training cycle, with experiments independently repeated 5 times to reduce randomness. The final results are presented as a mean-variance plot in Fig \ref{Fig:PPO_buffer}.

The average return over the last 20 episodes across five independent runs is 3829.04, used as the mean convergence value. While the Buffer-based PPO does not exceed the trajectory-based PPO in final scores, it markedly improves efficiency. For example, as shown in Table \ref{Tab:time_cost_cmp}, one run took 112 seconds for sampling and 54 seconds for updates, totaling 166 seconds—about 66 times faster than the trajectory-based approach. The update time proportion rose from 3.97\% to 32.53\%, showing effective balance between sampling and updates, thus boosting training efficiency.

\begin{table}[!htb]
	\centering
	\caption{Time cost comparison of PPO variants}
	\label{Tab:time_cost_cmp}
	\resizebox{0.45\textwidth}{!}{
		\begin{tabular}{@{} l cccc @{}}
			\toprule
			\textbf{Algorithm} & \textbf{Sampling} & \textbf{Policy Updates} & \textbf{Total} & \textbf{Update Ratio} \\
			\midrule
			Trajectory PPO          & 10,536s & 436s  & 10,972s & 3.97\%  \\
			Buffer PPO   & \bfseries 112s & \bfseries 54s & \bfseries 166s & \bfseries 32.53\% \\
			\bottomrule
	\end{tabular}}
\end{table}

As shown in Table \ref{Tab:step_cost_cmp}, from the perspective of training steps, the trajectory-based PPO algorithm utilized 69,046,726 steps, while the Buffer-based PPO required only 61,440 steps—a difference of approximately 1124 times. Despite this vast gap, the final training performance remains nearly identical. A test of 10,000 episodes was conducted using the better training result from one of the runs, yielding an average score of 4124.47, and the weights of this training result are shown in Table \ref{table:ppo_weight}.

\begin{table}[!htb]
	\centering
	\caption{Step cost comparison of PPO variants}
	\label{Tab:step_cost_cmp}
	\resizebox{0.45\textwidth}{!}{
		\begin{tabular}{@{} l c c @{}}
			\toprule
			\textbf{Algorithm} & \textbf{Total Steps} & \textbf{Avg. Converged Score} \\
			\midrule
			Trajectory PPO          & 69,046,726 & \bfseries 3,840.30 \\
			Buffer PPO (optimized)  & \bfseries 61,440   & 3,829.04 \\
			\bottomrule
	\end{tabular}}
\end{table}

\begin{table}[htb]
	\centering
	\caption{Best weights obtained from PPO training}
	\label{table:ppo_weight}
		\begin{tabular}{@{}lcc@{}}
			\toprule
			\textbf{Feature} & \textbf{Weight}  \\ \midrule
			Landing height & -0.51  \\
			Eroded piece cells & 0.16 \\
			Row transitions & -0.40 \\
			Column transitions & -0.75 \\
			Holes & -0.18 \\
			Board well & -0.39 \\
			Hole depth & -0.17 \\
			Rows with holes & -0.83 \\
			Diversity & 0.36 \\ \bottomrule
	\end{tabular}
\end{table}

This significant difference in training steps can be attributed to two phases:
\begin{enumerate}
	\item[(i)] \textbf{Early Training with Poor Sample Quality}: In Tetris, each piece is 4 cells, suggesting an average of 2.5 pieces per cleared row, yielding about 1 point per 2.5 steps. Early on, the agent's random action selection results in low scores (often single digits or zero), producing poor-quality samples with little value for policy optimization. The trajectory-based PPO, sampling full games, generates only tens of steps per game with minimal strategic insight, leading to ineffective updates. This necessitates repeated sampling of low-quality data, inflating step counts, as reflected by slow, low step growth in early training.
	\item[(ii)] \textbf{Late Training with Suboptimal Update Timing}: As training progresses, the agent's improved strategy yields higher scores (e.g., 1250 points over 4000-5000 steps). Valuable samples accumulate mid-game, but the trajectory-based PPO waits for game completion to update, wasting time on potentially suboptimal actions. In contrast, the Buffer-based PPO updates promptly after collecting sufficient samples, leveraging data more efficiently and reducing the step gap in later stages.
\end{enumerate}

\paragraph{\textbf{Generalization Verification}}

All experimental training in this paper is conducted based on a 10×10 mini-board and a random block generator, without designing an independent training process for the standard 10×20 board. During training on the 10×10 mini-board, the model gradually learns core operational logics including block movement, elimination judgment, and layout optimization. Such logics are partially decoupled from the board size, endowing the trained model with cross-size generalization and transfer capability that can be directly applied to the standard 10×20 board scenario. Table \ref{tab:ppo_scores_10x20} presents the test results on the standard board. Considering performance overhead, the experiment is run independently 100 times and the average value is taken.

\begin{table}[htbp]
    \centering
    \caption{Scores of two PPO algorithms on the 10$\times$20 game board (Mean(SD))}
    \label{tab:ppo_scores_10x20}
    \begin{tabular}{lc}
        \toprule
        Algorithm         & Score  \\
        \midrule
        Trajectory PPO    & 32,110,545.18(30,425,374.19)    \\
        Buffer PPO        & 13,936,917.72(13,198,381.24)    \\
        \bottomrule
    \end{tabular}
\end{table}

The results show that although the model can run stably and yield valid scores on the standard board, after direct cross-size transfer, its performance is significantly inferior to algorithms such as CBMPI compared with the training effect on the 10×10 board. This phenomenon mainly stems from two factors: first, the state space and the number of interactive steps per game of the 10×20 board increase significantly, imposing higher requirements on the model’s long-term planning ability and value estimation accuracy; second, the optimal layout strategy (e.g., compact stacking) for the 10×10 board cannot fully adapt to the vertical spatial characteristics of the standard board, limiting strategy applicability. Meanwhile, when extending the bitboard technique and Buffer mechanism proposed in this paper to the standard 10×20 board, two core challenges must be addressed: the significant increase in state space and interactive steps, which may greatly raise the average number of interactive steps per game, leading to distribution shift of Buffer samples and further affecting the stability of policy update; and value estimation bias caused by long-sequence decision-making. The longer game cycle of the standard board amplifies cumulative errors of long-term rewards, demanding higher accuracy in post-state value evaluation.

Moreover, to verify the adaptability of the proposed algorithm to different block generation rules, this experiment adopts three typical rules: Random, 7-Bag, and adversarial Z/S block sequences. The average score of the agent under each rule serves as the core evaluation metric. Random is used as the performance baseline, 7-Bag is the uniform random block generation rule commonly used in engineering, and the adversarial Z/S sequence is a deliberately designed harsh block generation rule to test the agent’s anti-interference ability and robustness. In the experiment, 10,000 independent test rounds are performed on the 10×10 board for each rule, and the average of multiple test results is taken as the final data to eliminate the impact of random errors on experimental results. Comparative experiments are completed by three agents under the same experimental environment, parameter configuration, and test rounds, with the results shown in Table \ref{tab:agent_scores_rules}.

\begin{table}[htb]
    \centering
    \caption{Scores of agents under different block generation rules (Mean(SD))}
    \label{tab:agent_scores_rules}
    \resizebox{0.45\textwidth}{!}{
        \begin{tabular}{lccc}
            \toprule
            Block Generation Rule & CBMPI      & Trajectory PPO                        & Buffer PPO                            \\
            \midrule
            Random (Baseline)     & 4,300      & \makecell{4,965.44 \\ (4884.14)}      & \makecell{3,590.90 \\ (3545.05)}      \\
            7-Bag                 & 251,501.20 & \makecell{173,489.37 \\ (177,212.68)} & \makecell{198,206.23 \\ (198,903.07)} \\
            Adversarial Sequence  & 75         & 34(0)                  & 53(0)                 \\
            \bottomrule
        \end{tabular}}
\end{table}

The results indicate that the performance of the three algorithms is highly consistent with changes in block generation rules: all agents achieve optimal performance under the widely used 7-Bag uniform rule in engineering; under the adversarial sequence rule, all agents drop to extremely low performance levels, showing sensitivity to harsh block sequences. All agents exhibit obvious performance shortcomings under the adversarial sequence rule, and robustness optimization research can be carried out in this direction in future work.

\section{Conclusion}
This paper proposes an efficient bitboard-based implementation scheme for Tetris, which significantly enhances the training capability of reinforcement learning agents for complex games. By redesigning the board and block structures with bitboards and leveraging bitwise operations to accelerate core processes (including block falling, collision detection, line clearing, and DT feature extraction), its performance is improved by 53 times compared to OpenAI Tetris-Gym. With the performance advantages of bitboards and Python-Java integration implemented via Jpype, rapid prototype development and experimental verification can be achieved. On this basis, this paper designs an afterstate-based evaluation Actor network and a buffer-optimized PPO algorithm, effectively addressing the unique challenges posed by the massive state space and long-term planning in Tetris. 

It should be noted that the method proposed in this paper does not take pursuing the absolute highest score as the sole objective, but focuses on achieving competitive policy performance at extremely low time and computing power costs. Experimental results show that the buffer-based PPO algorithm achieves an average score of 3829 on the 10$\times$10 board with only 61,440 interaction steps and approximately 3 minutes of training time. Although its absolute highest score is slightly lower than some state-of-the-art (SOTA) methods that rely on hundreds of millions of samples for training, its time cost advantage holds significant engineering application value. 

Through the combination of underlying bitboard acceleration and high-level buffer mechanisms, the framework in this paper enables a complete training process to be completed within a minute-level time scale, thereby significantly improving the practical value of Tetris as a rapid verification platform for reinforcement learning. Future research work can focus on the following directions to provide references for subsequent algorithm optimization: 
First, explore feature fusion strategies by combining traditional DT features with deep learning features, compare the performance differences of different feature combinations in state representation accuracy and training efficiency, and further enhance the model's ability to capture complex board states; 
Second, optimize the network structure design by attempting to introduce deep network structures such as MLP and Transformer, or integrate attention mechanisms, analyze the marginal gains of complex network structures on modeling accuracy in the massive state space of Tetris, and verify the adaptability of deep networks to the optimization framework proposed in this paper.

\section{Code Availability}
The source code for the bitboard-based Tetris AI is hosted on GitHub at: \url{https://github.com/GameAI-NJUPT/BitboardTetris}. All code is released under the MIT License, ensuring free access, modification, and distribution for academic and non-commercial purposes.

\begin{acks}
This study was supported by the National Natural Science Foundation,China (Nos. 62276142, 62206133, 62202240, and 62506172).
\end{acks}

\bibliography{references.bib}

\end{document}